\documentclass{article}

\usepackage[utf8]{inputenc} 
\usepackage[T1]{fontenc}    
\usepackage{hyperref}       
\usepackage{url}            
\usepackage{booktabs}       
\usepackage{amsfonts}       
\usepackage{amsmath}       
\usepackage{nicefrac}       
\usepackage{microtype}      
\usepackage{xcolor}         
\usepackage{graphicx}

\usepackage{amssymb}
\usepackage{mathtools}
\usepackage{amsthm}
\usepackage{algorithmic}
\usepackage{algorithm}

\usepackage{geometry}
\usepackage{subcaption}

\newtheorem{assumption}{Assumption}
\newtheorem{proposition}{Proposition}
\newtheorem{lemma}{Lemma}
\newtheorem{theorem}{Theorem}

\title{Practical Performative Policy Learning with Strategic Agents}

\author{
Qianyi Chen${}^1$, Ying Chen${}^2$, Bo Li${}^{1}$\thanks{Corresponding author}\\
${}^{1}$School of Economics and Management, Tsinghua University\\
${}^{2}$Department of IEOR, University of California, Berkeley\\
\texttt{cqy22@mails.tsinghua.edu.cn}\\
\texttt{ying-chen@berkeley.edu}\\
\texttt{libo@sem.tsinghua.edu.cn}\\
}

\begin{document}

\setlength{\parindent}{0pt}
\setlength{\parskip}{5.5pt}
\date{}
\maketitle

\begin{abstract}

This paper studies the performative policy learning problem, where agents adjust their features in response to a released policy to improve their potential outcomes, inducing an endogenous distribution shift. There has been a growing interest in training machine learning models in strategic environments, including strategic classification \cite{hardt2016strategic} and performative prediction \cite{perdomo2020performative}. However, existing approaches often rely on restrictive parametric assumptions: micro-level utility models in strategic classification and macro-level data distribution maps in performative prediction, severely limiting scalability and generalizability.
We approach this problem as a complex causal inference task, relaxing parametric assumptions on both micro-level agent behavior and macro-level data distribution. Leveraging bounded rationality, we uncover a practical low-dimensional structure in distribution shifts and construct an effective mediator in the causal path from the deployed model to the shifted data. We then propose a gradient-based policy optimization algorithm with a differentiable classifier serving as a substitute for the high-dimensional distribution map. Our algorithm efficiently utilizes batch feedback and limited manipulation patterns. Our approach achieves high sample efficiency compared to methods reliant on bandit feedback or zero-order optimization. We also provide theoretical guarantees for algorithmic convergence. Extensive and challenging experiments\footnote{The code is available at \url{https://github.com/Cqyiiii/Practical-Performative-Policy-Learning-PPPL}.} on high-dimensional settings demonstrate our method’s practical efficacy.

\end{abstract}

\section{Introduction}


The interaction between machine learning models and society has become an important topic, particularly due to the data distribution shifts induced by human behavior. The canonical learning paradigm, which assumes a static data distribution, often fails in strategic environments. A growing body of work addresses data distribution shifts arising from the strategic behavior of agents, which differ fundamentally from classical exogenous distribution shifts in their cause—these shifts are driven by the agents’ organic strategic responses to deployed algorithms in pursuit of better outcomes.

The first line of work is strategic classification with seminal papers \cite{bruckner2012static, hardt2016strategic}, where the decision process of an agent is usually formulated as utility function maximization. An agent therein seeks to balance the impact of the predictive algorithm under the hypothetical manipulated feature value and the cost of feature manipulation. The strategic classification paradigm essentially characterizes the micro-level structure of distribution shift as the best response of every specific agent given their original features, knowledge of the predictive model, and manipulation cost. There has been some recent development in strategic classification relaxing such assumptions \cite{ghalme2021dark, lechner2023unknown}, while practical model optimization, still relies on parametric specification of micro-level utility \cite{levanon2021practical}. 

Performative prediction \cite{perdomo2020performative} is a more general formulation, aiming to relax stringent agent utility modeling assumptions. From a causal perspective, performative prediction views the predictive model as a complex intervention on the data distribution. This perspective gives rise to a core concept, the distribution map $\mathcal{D}(\theta)$, which directly links the parameter $\theta$ of a predictive model and the shifted data distribution. There has been methods proposed to learn the performative optimal $\theta$ and the map $\mathcal{D}(\theta)$ such as bandit approach \cite{jagadeesan2022regret} and gradient-based optimization with parametric specifications on $\mathcal{D}(\theta)$ \cite{miller2021outside, izzo2021learn}, such as $\mathcal{N}(\mu(\theta),\sigma^2)$. For high-dimensional $\theta$ in practice, the former incurs bad sample efficiency and the latter incurs severe risk of macro-level data distribution misspecification. 


The primary objective of this paper is to make performative learning practical with mild structured assumptions. We first note that the model parameter $\theta$ is commonly high-dimensional in practice, which makes distribution map $\mathcal{D}(\theta)$ require a large number of algorithm deployments with different model parameters $\theta$ to generate sufficient variation. This requirement severely restricts the efficiency of learning. Second, at the conceptual level, we argue that it is not only inefficient but also incorrect to take the model parameter $\theta$ as the intervention instead of the predictive model $f_\theta$. For a neural network $f_\theta$, the parameter $\theta$ could be meaningless when separate from architecture, which makes the mapping from model parameter to distribution parameter very vague and hard to learn. In fact, the deployed model $f_\theta$ is the object that agents interact with, which is also reflected in the micro-level mechanism often used in strategic classification, e.g. refer to Equation (\ref{eq:strategic_classification}). These limitations hinder the practical implementation of existing performative prediction algorithms that rely on modeling $\mathcal{D}(\theta)$. Instead, our work first acknowledges $\mathcal{D}(f_\theta)$ and endeavors to leverage agent behavior structure in typical strategic environments to achieve greater dimension reduction of $\mathcal{D}(f_\theta)$.




To practically simplify the mapping $\mathcal{D}(f_{\theta})$, we draw motivation from the concept of bounded rationality, rooted in the behavioral economics literature \cite{conlisk1996bounded}. Bounded rationality suggests that agents may be unable to fully utilize all available information due to cognitive limitations. In our context, this means that an agent may lack the capacity to execute optimization procedures involving a highly complex prediction function $f_\theta$, even if the function is fully disclosed. Instead, the optimization they can realistically perform is \textbf{selecting the alternative with the highest utility from a finite set of options}.

This perspective is particularly relevant in performative learning, which involves a principal interacting with a population of agents—for instance, a bank handling numerous loan applicants or an e-commerce platform serving many consumers. This setting contrasts with the fully rational framework of classical game theory. For an agent with feature $x$, it is neither tractable nor necessary to exploit the entirety of the information contained in $f_\theta$. Instead, the influence of $f_\theta$ on the agent can be distilled into evaluations at specific feature values, e.g., $(f(x_i))_{i=1}^m$. Correspondingly, the agent’s consideration set comprises points $(x_i)_{i=1}^m$. We term this scenario limited manipulation and initially adopt it as an assumption for convenience. Nevertheless, in Section~\ref{sec:discretization}, we further demonstrate that this manipulation pattern can be \textbf{incentivized by a partially personalized policy}, which the principal can design.

\subsection{Our work}
In this paper, we build upon the literature on performative prediction and aim to operationalize the idea of taking the deployed model as the (functional) intervention imposed on data distribution, rather than the parameter of this model. To achieve this, we focus on settings with limited strategic behavior, where the complexity of the intervention is significantly reduced, making performative learning practical even for high-dimensional $\theta$. 

Our dimension reduction strategy further enables effective utilization of variation in the batch samples collected every round, since we choose to model the micro-level response of each agent. This contrasts with approaches that rely on a single bandit feedback per round of policy deployment, such as summarizing the observed data into a mean vector to estimate $\mu(\theta)$ in $\mathcal{N}(\mu(\theta), \sigma^2)$. We view performative learning as an extension of \textbf{supervised learning} rather than online learning and argue that methods relying on bandit feedback are highly inefficient in the performative learning context, particularly given the substantial costs associated with algorithm deployment.

We adopt principal-agent terminology and study how the principal can efficiently learn a policy for treatment allocation while incentivizing strategic agents. Specifically, we consider a scenario where agents modify some of their features through genuine effort rather than deception. This implies that feature manipulation causally affects the agent’s outcome and represents an abundant class of strategic behavior. When a well-crafted policy is released, agents are motivated to respond by exerting effort to change certain features for better outcomes, such as through a background boost. In contrast, if the policy is poorly designed, it may inadvertently encourage some agents to manipulate their features in directions undesirable to the principal.

Our study focuses on policy learning with binary treatment, as it represents a more general task targeting decision-making scenarios, compared to the binary classification task commonly studied in strategic classification and performative prediction. Another important motivation for this focus besides generality is that incentive-aware machine learning often involves interaction between deployed models and agents. These interactions are conducted by decisions rather than predictions, such as loan approval in loan applications or offers in college admissions—both classic examples in related works. Furthermore, we clarify that $f_\theta$ refers to the prediction function, while $\pi_\theta$ denotes the policy function throughout this paper. Nonetheless, both of them could serve as the intervention imposed on the data distribution.

As an alternative to strong parametric assumptions on data distribution, we propose to leverage the inherent structure in the feature space. We consider splitting the feature $x\in \mathcal{X}$ into discrete manipulatable feature $u\in \mathcal{U}$ and fixed features $v\in \mathcal{V}$. For example, in a college admissions scenario, the college acts as the principal, while the applicants are the agents. Most demographic attributes, such as name, gender, age, and school, are challenging to manipulate or change, especially given \textbf{an effective checking system}. However, these attributes are essential for colleges to determine treatment. There are also a few features such as GRE and TOEFL grades that can still be boosted by the efforts of the applicant during the application year. The varying emphasis colleges place on these scores—conveyed through official statements or third-party platforms \cite{ghalme2021dark}—influences how applicants allocate their efforts, often depending on their fixed features. Motivated by this example, we further consider the scenario where $\dim \mathcal{U} \ll \dim \mathcal{V}$ in this paper.


This regime offers substantial benefits by significantly reducing the dimensionality of the complex intervention $\pi_\theta$. To summarize the impact of $\pi_\theta$ on a specific agent with fixed features $v$, it suffices to only consider the evaluation vector $(\pi_\theta(u, v))_{u \in \mathcal{U}}$. In many practical scenarios, $u$ can be viewed as a feature type with cardinality much smaller than the dimension of $\theta$. Consequently, shifting the focus from $\theta$ to $\pi_\theta$ as the intervention often facilitates considerable dimensionality reduction. This perspective, along with the resulting dimension reduction, is highly nontrivial, as it enables efficient learning under far more relaxed assumptions compared to existing works.

To support gradient-based optimization for performative learning, we propose modeling the decision outcome directly, bypassing the agent’s decision process, which typically relies on a parametric micro-level mechanism, such as utility maximization. The decision outcome can be characterized as the distribution of the manipulatable feature $u^\prime$ reported by the agent, given the limited information about the decision function that would be used by this specific agent. Specifically, we choose to model $p(u^\prime \mid (\pi_\theta(u,v))_{u\in\mathcal{U}})$, which we refer to as the behavior model. We argue that the original feature, prior to manipulation, is not only difficult to acquire in practice but also unnecessary in our scenario—it can be treated as hidden heterogeneity.

To delineate, this behavior model serves as a substitute for the true distribution map $\mathcal{D}(\pi_\theta)$ and enables gradient-based optimization by providing an estimate of the gradient $\partial \mathcal{D}(\pi_\theta) /\partial \theta$. In practice, we find that both the neural network and Gaussian process classifier are qualified to serve as this behavior model. Moreover, based on this agent behavior model, we are no longer confined to the traditional scenario of deterministic manipulation, broadening its applicability to more nuanced micro-level settings such as private types \cite{levanon2022generalized} and random or noisy responses \cite{jagadeesan2021alternative}.

In addition, we establish the convergence guarantee for our gradient-based policy optimization algorithm through a novel approach: we rely on the realizability assumption in a reproducing kernel Hilbert space (RKHS), as opposed to the parametric assumption on distribution and finite difference method used in algorithm~\cite{izzo2021learn}, both of which becomes highly vulnerable when the dimensionality of data increases. Even though the true distribution map induced by best response mechanism often admits discontinuities~\cite{levanon2021practical} and the realizability assumption is violated therein, we demonstrate that our method still achieves stable convergence and exhibits excellent performance, in the simulation study.


\textbf{Contributions.} Our contributions are summarized as follows.
\begin{enumerate}
    \item We investigate a novel and \textbf{practical} setting for performative policy learning, characterized by limited manipulation driven by bounded rationality, which captures a wide range of realistic scenarios. Our formulation relaxes traditional strong parametric assumptions while maintaining the feasibility of performative learning.
    
    \item We propose directly learning the agent’s decision outcomes, bypassing the decision-making process. This approach eliminates the need for parametric utility specifications while offering an effective \textbf{differentiable} substitute for the distribution map.

    \item We develop a gradient-based algorithm for performative policy optimization that \textbf{scales} efficiently with the dimensionality of both data distributions and policy parameters. Additionally, we establish a solid convergence guarantee through a novel approach.
    
    \item We conduct extensive and challenging experiments in synthetic and semi-synthetic settings with \textbf{high-dimensional} model and data to demonstrate the advantages of the proposed method compared to classic baselines. These experiments further validate the significant performance improvements brought by our conceptual contributions.

\end{enumerate}

\subsection{Related works} 

\textbf{Strategic Classification.} The basic target of this area is training a classifier that is robust to manipulation. As discussed previously, there are many limits on the original setting proposed in \cite{bruckner2012static, hardt2016strategic}. A strand of recent research makes efforts on multiple aspects to relax the original assumptions, especially the precisely specified utility and deterministic best response \cite{dong2018strategic, ghalme2021dark, jagadeesan2021alternative, chen2020deltaball, lechner2023unknown, rosenfeld2024oneshot, shao2024strategic}, in which agent admits bounded manipulation and/or principal has access to partial information. 
Specifically, \cite{bjorkegren2020manipulation} considers estimating the post-manipulation feature distribution, though the discussion there heavily relies on linear decision rule and quadratic manipulation cost. Another strand of work focuses on the causal mechanism in the strategic environment, where the outcome of the agent is causally affected by feature manipulation. An important topic wherein is incentivizing improvement \cite{kleinberg2020classifiers, shavit2020causal, bechavod2021gaming, haghtalab2021maximizing, chen2023learning}. In addition, \cite{miller2020strategic} characterizes incentivizing improvement as a non-trivial causal inference problem. \cite{DBLP:conf/icml/HorowitzR23} emphasizes the prediction accuracy and examines the interaction of feature shift and outcome shift from a causal perspective. Specifically, both papers leverage the split of features into causal and non-causal ones, which resonates with our idea to exploit the structure of feature space. Furthermore, very few papers have addressed efficient model optimization in the area of strategic classification, since differentiating through best-response is non-trivial even when agent utility is given. \cite{levanon2021practical} propose to utilize the convex-concave procedure to make gradient-based optimization feasible, though parametric utility specification is still inevitable there.



\noindent\textbf{Performative Prediction.} The seminal paper \cite{perdomo2020performative} proposes a series of pivotal concepts including distribution map $\mathcal{D}(\theta)$, performative risk ${\mathbb{E}}_{z \sim \mathcal{D}(\theta)} \ell(z ; \theta)$, where $\ell$ denotes the loss function. Based on this framework, they propose two types of solutions, namely performative stable and optimal solutions, where the former is the fixed point of performative risk and the latter is the minimizer of performative risk. They also propose a repeated training strategy for finding the performative stable solution. Other papers targeting performative stable solutions include \cite{mendler2020stochastic, drusvyatskiy2023stochastic, mofakhami2023ppnn}. Nonetheless, a possible large gap between performative stable and optimal solution has been pointed out by \cite{miller2021outside}, which motivates the pursuit of the performative optimal solution by directly minimizing the performative risk. Generally, to minimize the performative risk through gradient-based optimization, it can not be steered around to model the distribution map $\mathcal{D}(\theta)$. In this regard, \cite{izzo2021learn} proposes to approximate the gradient of the data distribution parameter concerning the model parameter, called performative gradient, see also \cite{lin2024plug, wenjing2023zero}. This line of work heavily relies on macro-level parametric assumptions of data distribution. This formulation also leads to data scarcity as each round of model deployment only gives a single feedback, called bandit feedback~\cite{jagadeesan2022regret}. These two drawbacks in existing literature motivate us to incorporate micro-level behavior structure exploitation to reduce complexity and increase the richness of feedback. Additionally, \cite{mendler2022anticipating} studies the scenario that predictions only influence outcomes and formulates the idea of prediction as intervention and examines the identification of shift of outcome $Y$ due to revealing prediction $\hat Y$. The idea of taking prediction as the intervention inspires us to reduce the dimension of intervention from a causal mechanism perspective. \cite{kim2023making} also studies this scenario and illustrates the complexity of finding performative optimal classifiers is substantially reduced.

\noindent\textbf{Policy Learning with Distribution Shift.} Policy learning algorithms typically leverage counterfactual prediction to prescribe decisions with good value performance. In general, policy learning is tightly related to the classification problem \cite{irpan2019off}. The optimal policy in a static setting admits a simple cutoff structure, namely, we only need to allocate treatment to those with positive conditional average treatment effect (CATE) \cite{manski2004statistical, athey2021policy}. Beyond the basic setting, decision making in non-stationary environment is a long-standing topic \cite{besbes2015non, cheung2022hedging}. A series of recent papers also propose to use distributionally robust optimization (DRO) \cite{rahimian2019distributionally, besbes2022beyond, gao2024wasserstein} for policy learning that is robust to either marginal feature shift or joint distribution shift \cite{mo2021learning, kallus2021minimax, kallus2021more, kallus2022doubly, adjaho2022externally}. Previous studies on distribution shifts often focus on shifts that are exogenous to the deployed policy, rather than those arising from strategic responses and performativity driven by the deployed policy.
Roughly speaking, performative policy learning can be seen as directionally adaptive to specific structured distribution shifts, rather than agnostically robust within a distributional ball. For further comparison, see \cite{munro2024treatment}. More relevant to our paper, \cite{munro2024treatment} considers policy learning problems with possible distribution shifts induced by agents' strategic behaviors. He views the policy function $\pi$ as an intervention and carries out functional analysis for the gradient of policy value concerning the policy function. He shows the structure of optimal policy may no longer be in a deterministic cutoff form, which motivates us to develop a practical policy learning algorithm that extends beyond CATE estimation. Since no further complexity reduction is considered in \cite{munro2024treatment}, the policy optimization therein can not scale up beyond the one-dimensional feature. \cite{tsirtsis2024optimal} studies policy learning with purely discrete features and structured knowledge on manipulation cost, while no practical learning algorithm is discussed there. \cite{sahoo2024policy} addresses policy learning with competing agents and carries out gradient-based policy optimization through the specification of agent utility with noisy manipulation. \cite{liu2024contextual} studies the problem of strategic contextual pricing, where a micro-level utility is specified with unknown parameters. They implement a two-phase algorithm combined with the doubling trick, which first collects data with a uniform policy, and then optimizes the policy only once for future deployments. Though lack of refined design on policy optimization, an optimal regret rate is achieved there.


\section{Problem Setup}

\subsection{Multi-period setting}

We will adopt Neyman-Rubin potential outcome notation \cite{rubin1974estimating} for causal effect estimation and policy learning. In addition, a causal graph will also be employed for better illustration in the later section. We define our problem as multi-period interactions between a principal and a population of agents. For epoch $t=1,2,\dots, T$, we consider following event flow: 
\begin{enumerate}
    \item Principal releases policy $\pi=\pi_{\theta_{t}}$
    \item $n$ new agents arrive, and make decisions on feature manipulation. This generates the post-manipulation feature  $X_i^\pi = x_i$ and potential outcomes $Y_i(1), Y_i(0)$, for $i=1,2,...,n$.
    \item Agents report the post-manipulation features.
    \item Principal observes the $x_i$ and allocate binary treatment $Z_i$ with treatment probability $\pi_{\theta_t}(x_i)$. 
    \item Agent $i$ observes the binary treatment $Z_i$.
    \item Principal observes the outcome $Y_i(Z_i)$
    \item Principal updates the parameter $\theta_t$ 
\end{enumerate}
We assume the principal would release the policy $\pi$, which may deviate from practical scenarios. Nevertheless, there are third-party consulting agencies or crowdsource platforms that collect historical data constantly and implement reverse engineering to dissect the principal's decision rule. Furthermore, the policy usually will not keep changing drastically, which means that reverse engineering is feasible. On the other hand, the release of policy can often be used as a strategy for incentivizing improvement (e.g. the college wants to incentivize the applicants to refine their language skills to enhance performance in their future studies). At last, we clarify that $Y_i(1)=Y_i(1;X_i^\pi), Y_i(0)=Y_i(0;X_i^\pi)$, while we suppress $X_i^\pi$ here for neatness. 

At each epoch, new agents arrive, with their features sampled i.i.d. from a fixed data distribution. Upon being informed of the policy $\pi_\theta$, agents exert genuine effort to modify part of their features and submit their final feature vector $x_i$. In this paper, we focus on scenarios where agents' feature manipulation causally influences their potential outcomes. Consequently, a well-designed policy can incentivize agents to change their features in the desired direction of principal, which generally corresponds to better outcomes $Y$. The principal’s decision then involves allocating a binary treatment $Z_i$, determined by the propensity $\pi_{\theta_t}(x_i)$. Following this, the principal observes the outcome of interest, which guides the update of $\theta_t$.


For illustration, consider the example of a loan application, where the principal is the bank and the agents are the loan applicants. The treatment in this context is the offer decision, and the potential outcome is the final revenue from the transaction, where $Y(0)=0$ and $Y(1)$ represents the expected return given a certain probability of default. This default probability can be causally influenced by the manipulation of specific features, such as the liquidity or risk level of applicant's assets. Agents can adjust their asset allocation—e.g., selling high-stakes assets like stocks and increasing the proportion of highly liquid assets such as cash—to improve their ability to fulfill the loan. The principal's objective is naturally to incentivize agents to enhance their ability to repay and reduce the default probability. However, a poorly designed policy may inadvertently motivate agents to increase their allocation to high-stakes assets, thereby raising the default probability.

\subsection{Gradient-based performative policy learning}

Now we can discuss the principal's learning problem. We define the CATE as the expected difference between treated and control outcomes for an individual given its submitted feature $x$.
\begin{equation}
    \tau(x) = \mathbb{E}[Y_i(1)-Y_i(0)\mid X_i^\pi=x]
\end{equation}
It is important to clarify that though we continue to use the notation $X_i^\pi$ to represent the feature influenced by policy $\pi$, the CATE remains indifferent to the specifics of $\pi$ once the submitted feature $x$ is determined. 

We further denote the probability density of $x$ after manipulation by $p(x;\pi_\theta)$, referred to as the performative distribution, which captures the influence of the policy $\pi_\theta$ on the data distribution. In this setting, the policy value can be expressed as:
\begin{equation}
    V(\pi_\theta)=\int \pi_\theta(x) \tau(x)p(x;\pi_\theta) \mathrm{d}x =\mathbb{E}_{x\sim p(\cdot;\pi_\theta)}[\pi_\theta(x) \tau(x)]
\end{equation}

Naturally, the principal's objective is to identify an optimal policy $\pi^*$ that maximizes the performative policy value:
\begin{equation}
    \pi^* = \operatorname{argmax}_\theta V(\pi_\theta)
\end{equation}

In this paper, we approach this problem through the lens of supervised learning. This perspective is particularly motivated by our goal of leveraging data collected in each interaction round as batch feedback, rather than relying on bandit feedback. Nevertheless, we clarify that, in the terminology of online learning, our problem concerns best-arm identification rather than minimizing the cumulative regret.

Next, we can derive the performative policy gradient as follows:
\begin{equation}\label{eq:true_policy_gradient}
    \nabla_\theta V(\pi_\theta) = \mathbb{E}_{x\sim p(\cdot;\pi_\theta)}[\nabla_\theta\pi_\theta(x) \tau(x)+\pi_\theta(x) \tau(x)\nabla_\theta\log p(x;\pi_\theta)]
\end{equation} 
The decomposition of the gradient into two distinct components is a key insight: one component is the gradient through the policy network $\pi_\theta$, and the other is through the density $p(x;\pi_\theta)$. This separation aligns with the concept of the performative gradient as discussed in \cite{izzo2021learn}. 

This equation is a direct result of applying REINFORCE estimator~\cite{williams1992simple} so we treat it as basics. We note that the critical part is the \textbf{formulation and evaluation of the gradient} $\nabla_\theta\log p(x_i;\pi_{\theta})$, where $\theta$ can be high-dimensional in practice, rendering past methods ineffective. Instead, we endeavor to model and estimate this gradient in a fully non-parametric way with practical assumptions.



\section{Main Methodology}

\subsection{Partition the feature space}
At the outset of this section, we outline our strategy for partitioning the feature space $\mathcal{X}$. We distinguish between manipulatable feature $u\in \mathcal{U}$ and fixed features $v\in \mathcal{V}$. We consider $\mathcal{U}$ as a discrete agent type space, which could represent primitive agent types or result from the discretization of continuous features. This aligns the bounded rationality in the agent's manipulation decision.

\begin{assumption}\label{assumption1}
    The feature $x\in\mathcal{X}$ can be separated to discrete manipulatable feature $u\in \mathcal{U}$ and fixed feature $v\in \mathcal{V}$. Specifically, $\mathcal{U}$ is a finite set.
\end{assumption}

For clarity, we defer the discussion of Assumption~\ref{assumption1} until Section~\ref{sec:discrete} due to the technical complexities involved. We now revisit the example of college admission for illustration. It is natural to assume that the policy changes each year, and that most of the applicant features are fixed, unable to be altered through efforts within the limited time before the application. For instance, features such as undergraduate school, major, GPA in the freshman and sophomore years, and demographics are typically beyond the applicant's control. A similar discussion applies to the scenario of a loan application, provided there is an effective checking system in place. Therefore, it is reasonable to assume that the dimensionality of $\mathcal{U}$ is much smaller than that of $\mathcal{X}$, which alleviates the challenges associated with discretization.

Finally, we note that such an assumption regarding the partition of the feature space with limited manipulation would become questionable in the context of classic strategic classification, where feature manipulation corresponds to cheating or gaming, rather than improving that is discussed in this paper. In practice, as the principal, we may only maintain a belief about which features are more sensitive or likely to be manipulated, with each feature having a certain probability of being faked. Nonetheless, we believe this assumption is reasonable in our context, as we are not concerned with the authenticity of the materials submitted.

\begin{figure}[t]
    \centering
    \begin{minipage}{0.48\textwidth}
      \centering
      \includegraphics[width=\linewidth]{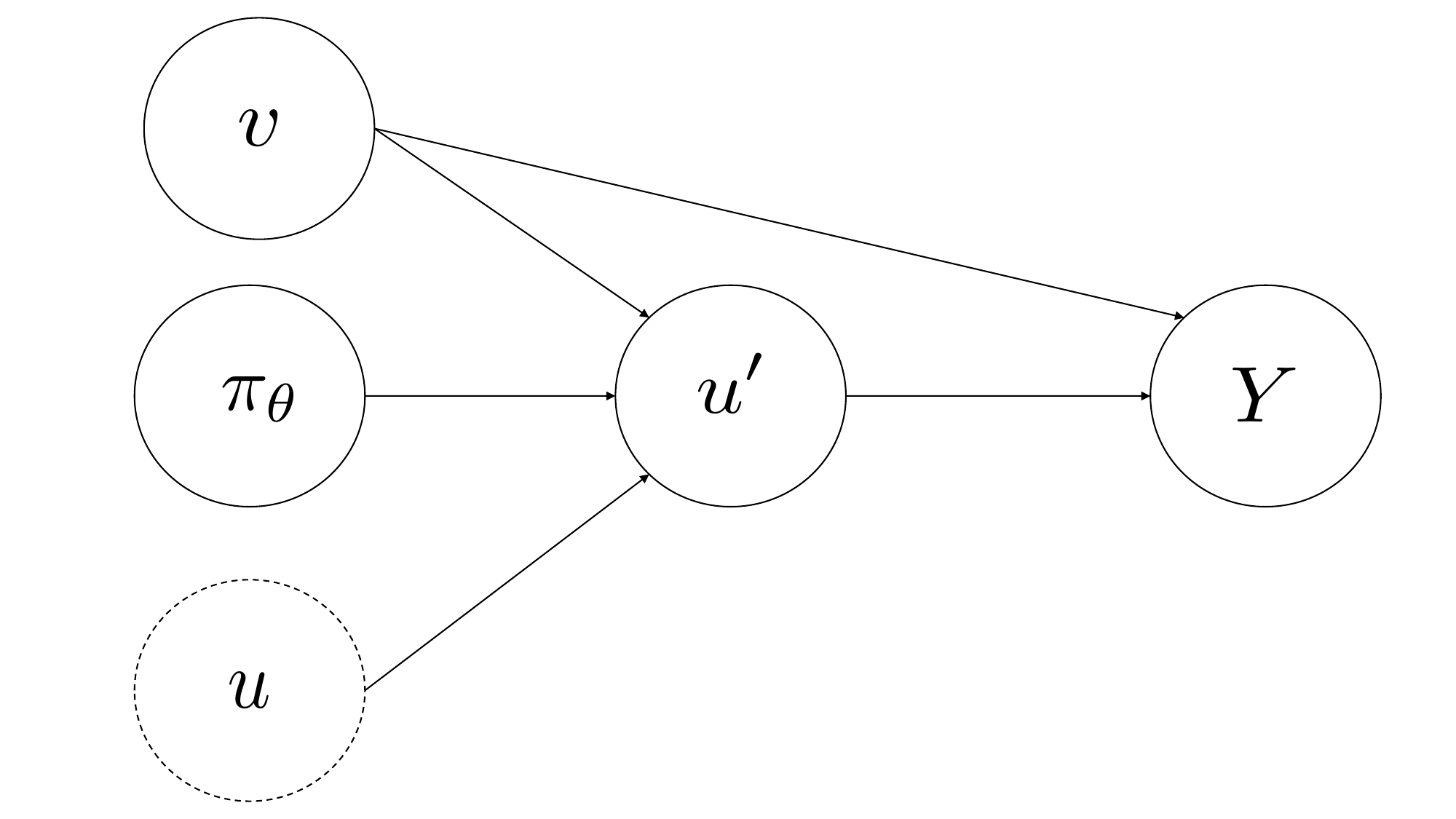}
      \subcaption{Default case}
      \label{fig:causal_graph_a}
    \end{minipage}\hfill
    \begin{minipage}{0.48\textwidth}
      \centering
      \includegraphics[width=\linewidth]{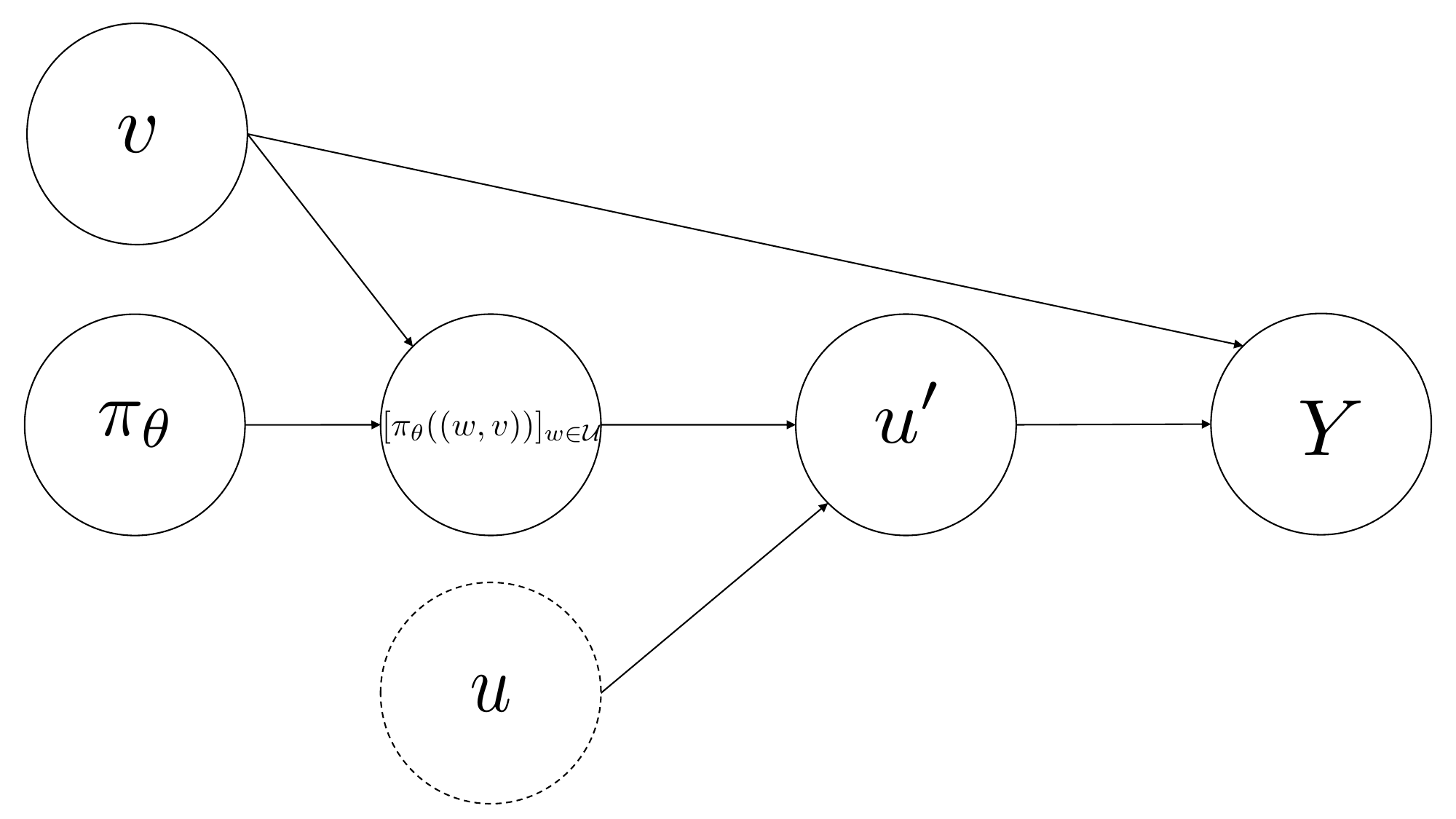}
      \subcaption{With proposed mediator}
      \label{fig:causal_graph_b}
    \end{minipage}
    \caption{The left causal graph is concerning features $u$, $v$, policy $\pi_\theta$, post-manipulation feature $u^\prime$, and outcome $Y$. Here the original manipulatable feature $u$ is unobserved and denoted by dashed circle. The main complexity lies in policy function $\pi_\theta$ acting as intervention. With the proposed partition of feature space, we argue that the policy evaluation vector $[\pi_\theta((w,v))]_{w\in \mathcal{U}}$ can act as mediator that summarizes the impact of policy $\pi_\theta$, as shown in the right graph.}
    \label{fig:causalgraph}
    \vskip -0.1in
\end{figure}


\subsection{Encapsulate the impact with evaluation vector}\label{sec:3.2}

Given this partition, we propose a method to capture the influence of the policy function $\pi_\theta$ on an individual agent $i$. This is achieved by evaluating the policy function across all values of the manipulable feature $u$, while keeping $v_i$ fixed, resulting in a vector $[\pi_{\theta}((u,v_i))]{u\in \mathcal{U}}$. This vector encapsulates all the information available to agent $i$ with the fixed feature $v_i$. The evaluations of the policy function $\pi_\theta$ serve as mediators in the causal path to the reported signal $X_i^\pi$, i.e., $(u_i^\prime,v_i)$, as illustrated in Figure~\ref{fig:causalgraph}.

\begin{assumption}[Causal mechanism]\label{assumption2} 
We assume the manipulatable feature $u^\prime$ satisfying:
\begin{equation}
    u^\prime = u^\prime(u,\pi_\theta(\cdot,v))
\end{equation}
This implies the following two points:
\begin{enumerate}    
    \item An agent with feature $(u,v)$ will not consider the treatment probability assigned to cases with a different fixed feature $v' \neq v$ when making the decision to manipulate the feature.
    \item The fixed feature $v$ will influence the agent's decision on $u'$ only through $\pi_\theta(\cdot,v)$.
\end{enumerate}   
\end{assumption}

This mechanism restricts the view of the agent to its achievable utilities when it decides how to manipulate, and excludes some complex cases such as cross-agent interference. For further motivating this assumption, let us examine the classic micro mechanism in the strategic classification:
\begin{equation}\label{eq:strategic_classification}
    u^\prime = \operatorname{argmax}_{u} \pi_\theta(u,v)-c\|u-u_0\|_2
\end{equation}
here $u_0$ represents the original manipulatable feature, and $c$ is the coefficient of manipulation cost. The agent's utility is the probability of receiving treatment (or probability of being classified into the positive class in the classification task) minus the cost of feature manipulation. We can easily examine that the two points in Assumption~\ref{assumption2} are satisfied here. However, to avoid such a parametric specification, we choose to \textbf{abstract the basic information mechanism} and drop the parametric structure. 

\textbf{Remark}. When the second point in Assumption~\ref{assumption2} does not hold—that is, if there exists an edge from $v$ to $u^\prime$ in the causal graph—we would encounter an identification issue in uniquely determining the influence of evaluation vector $\left[\pi_\theta\left(\left(w, v_i\right)\right)\right]_{w \in \mathcal{U}}$ on $u^\prime$. This situation is similar to the case described in~\cite{mendler2022anticipating}. A solution to this issue, without imposing parametric assumptions on the true causal mechanism, is introducing exogenous variation into the evaluation vector. Rather than directly adding noise to $\pi_\theta$, here we can resort to the natural variation of $\theta$ introduced by multiple paths with random initialization and stochastic gradient ascent\footnote{The variation of $\theta_t$ arises from random initialization and samples observed before $t$-th epoch. Therefore, it is exogenous for the sample observed at the $t$-th epoch with fixed feature $v_i$, as we assume the samples are generated independently.} in the process of updating $\theta$. Nonetheless, we construct Assumption~\ref{assumption2} since the corresponding utility maximization process and its variants are widely considered in the related literature.

To clarify, we will denote the reported manipulatable feature by $u$ henceforth, as the principal cannot observe the original manipulatable feature and is not concerned with it. When there is potential ambiguity, we will specifically use $u^\prime$ to indicate the reported manipulatable feature, and $u_0$ to represent the original one, as shown in Equation (\ref{eq:strategic_classification}).

Now we combine Assumption~\ref{assumption1} and~\ref{assumption2} and get the following proposition:
\begin{proposition}\label{prop1}
We can simplify the performative data distribution
\begin{equation}\label{eq:prob_decompose}
\begin{aligned}
p(x;\pi_{\theta})=p(u\mid \zeta(v, \pi_\theta)) p(v)\\
\end{aligned}
\end{equation}
and subsequently transform the performative gradient
\begin{equation}
    \nabla_\theta \log p(x;\pi_{\theta}) = \nabla_\zeta \log p(u\mid \zeta) \nabla_\theta\zeta(v, \pi_\theta)
\end{equation}
Hereafter, we use $\zeta(v, \pi_\theta)$ to denote function evaluation vector $[\pi_{\theta}((w,v))]_{w\in \mathcal{U}}$.
\end{proposition}
The proof of Proposition \ref{prop1} is provided in Appendix \ref{app:proof_of_prop1}. In the following, we discuss the implications of the transformation introduced in this proposition, as well as the benefits it brings to the sample efficiency of performative learning.

In the most general case, modeling the performative distribution $p(x;\pi_\theta)$ involves characterizing the influence of a functional treatment $\pi_\theta$ on the distribution of the features $(u, v)$. Notably, due to the considered causal mechanism, we avoid the complexities associated with $v$ and $\theta$, both of which can be high-dimensional in practice. Instead, the intervention received by the agent is represented by $\zeta$ rather than $\pi_\theta$ or $\theta$. Furthermore, we do not need to model the distribution $p(v)$ if our sole objective is to optimize the policy $\pi_\theta$.

Additionally, it is worth noting that the end-to-end approach based on the concept of the distribution map $\mathcal{D}(\theta)$ would correspond to modeling the performative distribution $p(u \mid v, \pi_\theta)$ with $p(u \mid v, \theta)$, in contrast to $p(u \mid \zeta(v, \pi_\theta))$ in our methodology. We argue that this approach is flawed not only because $\theta$ is high-dimensional but also due to the conceptual understanding that $\pi_\theta$ serves as the true intervention on the data distribution, rather than $\theta$. In comparison, our method can be seen as a more reasonable means of elucidating the causal mechanism, thereby making performative policy learning practical.

Empirically, learning the conditional distribution $p(u \mid v, \theta)$ presents significant challenges, even in basic settings, as demonstrated by our experiments. This difficulty arises not only from the ambiguous relationship between $\theta$ and feature manipulation but also from the limited variation of $\theta$; specifically, the number of distinct $\theta$ values in the dataset equals the number of epochs $T$, which is much smaller than the total sample size $Tn$. This limitation underscores the necessity of leveraging batch feedback from a \textbf{supervised learning perspective} in performative learning, rather than relying on bandit feedback. Furthermore, it reinforces one of our central arguments that the true intervention is the function $\pi_\theta$, not the parameter $\theta$. Moreover, it highlights an implicit drawback in the notation of the distribution map $\mathcal{D}(\theta)$.

We now introduce another critical component of our methodology: utilizing a separate differentiable classifier, denoted as $h_\gamma$, to approximate the conditional distribution $p(u \mid \zeta(v, \pi_\theta))$. This approximation corresponds to the distribution $q(u \mid \zeta(v, \pi_\theta))$. 

To train $h_\gamma$, we use the observed feature $u_i$ as the supervision signal during each epoch $t$, framing the task as a (multi)classification problem with cross-entropy as the loss function. Specifically, $h_\gamma$ predicts the observed feature $u_i$ given $\zeta(v_i, \theta_t)$. Since this classifier encapsulates the agent's behavior—whether through a deterministic best response based on utility or a random response—we refer to $h_\gamma$ as our behavior model. The key idea is to bypass the complexity of the decision-making process by directly modeling the agents' decision outcomes. In practice, both neural networks and Gaussian process classifiers are well-suited to serve as the behavior model in this context.

\subsection{Strategic policy gradient}\label{sec:spg3.3}


Besides $h_\gamma$ described in the previous subsection, an estimator for the CATE is required. To this end, we adopt a meta-learner approach \cite{kunzel2019metalearners} to construct the estimator $\hat{\tau}(x)$. The resulting estimator for the true performative policy gradient,  as expressed in Equation (\ref{eq:true_policy_gradient}), is defined as follows:
\begin{equation}\label{eq:full_grad}
\hat g = \frac{1}{n}\sum_{i=1}^{n}[\nabla_\theta\pi_\theta(x_i) \hat\tau(x_i)
    +\pi_\theta(x_i) \hat\tau(x_i)\nabla_\theta
    \log( q(u_i\mid \zeta(v_i,\pi_\theta) )
]
\end{equation}
We denote the gradient estimator at epoch $t$ as $\hat{g}_t$. The learning algorithm, formally presented as Algorithm~\ref{alg:mmgd}, is referred to as \textit{Strategic Policy Gradient}. This method employs a policy gradient approach with an estimate of the true performative gradient, leveraging a behavior model that captures the outcomes of agents' strategic behavior. By doing so, it facilitates gradient-based policy optimization without imposing parametric assumptions on either the micro-level utility or the macro-level distribution.

We begin by adopting a warm-up stage to collect data for learning the behavior model $h_\gamma$, the CATE estimator $\hat{\tau}(x)$, and the policy $\pi_\theta$ in a preliminary manner. During this warm-up phase, we perform repeated gradient ascent using a vanilla gradient that does not account for the performative component. It is important to note that, at this stage, we will have $nT_0$ samples corresponding to $T_0$ distinct values of $\theta_t$, providing sufficient variation in the evaluation vector $\zeta$ for efficient learning of the behavior model $h_\gamma$. Once this phase is completed, we proceed with applying the full gradient estimate as presented in Equation (\ref{eq:full_grad}).

\begin{algorithm}[tb]
   \caption{Strategic Policy Gradient}
   \label{alg:mmgd}
\begin{algorithmic}[1]
   \STATE {\bfseries Input:} Time horizon $T$, warm up rounds $T_0$, learning rate $\eta_1,\eta_2$
   \STATE // Warm up stage, update $\theta_t$ with vanilla gradient
   \FOR{$t=1$ {\bfseries to} $T_0$}
   \STATE Deploy policy $\pi_{\theta_t}$
   \STATE $\mathcal{D}_t\leftarrow\{(x_i, z_i, Y_i(z_i), [\pi_{\theta_t}((u,v_i))]_{u\in \mathcal{U}})\}_{i=1}^n$
   \STATE Update CATE estimator $\hat\tau$
   \STATE $\theta_{t+1}\leftarrow \theta_t + \frac{\eta_1}{n}\sum_{i=1}^{n}\nabla_{\theta}\pi_{\theta_t}(x_i) \hat\tau(x_i)$
   \ENDFOR
   \STATE // Merge data collected in warm-up stage 
   \STATE $\mathcal{D}_{\text{warm}} \leftarrow \{\mathcal{D}_t\}_{t=1}^{T_0}$
   \STATE Train $h_\gamma$ on $\mathcal{D}_{\text{warm}}$    
   \STATE // Update $\theta_t$ with full gradient
   \FOR{$t=T_0+1$ {\bfseries to} $T$} 
   \STATE Deploy policy $\pi_{\theta_t}$
   \STATE $\theta_{t+1}\leftarrow \theta_t + \eta_2 \hat g_t$
   \ENDFOR
\end{algorithmic}
\end{algorithm}

Next, we provide insights to facilitate the successful implementation of our algorithm. These considerations also help elucidate the algorithm’s workings and the influence of performativity on policy optimization.

A critical component in modeling the performative effect is the behavior model, which we propose to parameterize using a neural network trained via a classification task. The granularity of discretization, corresponding to the number of classes in this task, should be carefully chosen based on the available sample size per epoch and the frequency of model deployments. Specifically, we empirically recommend setting $|\mathcal{U}|\in[5,20]$. To further explore this aspect, we analyze the impact of varying discretization granularities in our synthetic experiment.


Second, since the policy optimization problem can be highly non-convex, optimizers with momentum are a desirable option. However, as the data distribution echoes the deployed policy constantly, we argue that first-order momentum can degrade both the performance and stability of our algorithm once the warm-up stage concludes and the performative gradient estimate is applied. This is because gradients from earlier steps may no longer be relevant and could even mislead the optimization process. To address this challenge, we recommend using Adagrad~\cite{JMLR:v12:duchi11a}, which relies only on second-order momentum to adaptively normalize the gradient. Empirically, it performs very well in our experiments.

\section{Theoretical Analysis}

\subsection{Overview}


In this section, we establish a convergence guarantee for our method, the strategic policy gradient. 
We begin by addressing the complexities inherent in providing such a theoretical guarantee. The convergence of repeated retraining has been studied in the seminal work~\cite{perdomo2020performative}, which focuses on identifying a performative stable solution by framing it as a fixed-point problem. However, when the objective shifts to finding the performative optimal solution, ensuring convergence for gradient-based methods presents significantly greater challenges.
We argue that the key component lies in bounding the gap between the true performative gradient and its estimated counterpart by estimation error of the performative distribution, since the later one is the aspect that we can directly control.

In~\cite{izzo2021learn}, the authors establish a convergence guarantee for their proposed performative gradient descent algorithm. Their method assumes a parametric distribution family, specifically the Gaussian distribution, with performativity restricted to the mean vector. To address the critical component mentioned above, they utilize finite differences to construct the performative gradient estimate, relying on the estimation of the performative distribution. However, this approach suffers from poor scalability as data dimensionality increases.

In contrast, we propose a differentiable classifier, referred to as the behavior model, to estimate the performative distribution in a non-parametric manner. While $\mathcal{U}$ is discretized, directly bounding the gap between $\nabla_\theta p(u \mid \zeta(v, \pi_\theta))$ and $\nabla_\theta q(u \mid \zeta(v, \pi_\theta))$ remains challenging. However, we observe that if both the true performative distribution $p(u \mid \zeta(v, \pi_\theta))$ and its estimate $q(u \mid \zeta(v, \pi_\theta))$ lie within a common RKHS for all $u \in \mathcal{U}$, it becomes possible to establish a desirable bound. To clarify, the distribution functions $p$ and $q$ are defined as mappings $p, q: [0,1]^{|\mathcal{U}|} \rightarrow [0,1]$, for each $u \in \mathcal{U}$.

We now define $\Omega = [0,1]^{|\mathcal{U}|}$ as the space of evaluation vector $\zeta$. We proceed by introducing the following lemma:
\begin{lemma}\label{lemma1}
Given that the functions $f(\zeta)$ and $g(\zeta)$ both lie in an RKHS $\mathcal{H}$ with kernel function $K \in C^{2s}(\Omega \times \Omega)$ and $s \geq 2$, we claim that the gradients $\nabla_\zeta f(\zeta)$ and $\nabla_\zeta g(\zeta)$ exist. Furthermore, we have the following bound: 
\begin{equation}
    \|\nabla_\zeta f(\zeta)-\nabla_\zeta g(\zeta)\| \leq\|f-g\|_{\mathcal{H}} \cdot \left\|\left( \left\|\nabla_{\zeta_i} K(\zeta, \cdot)\right\|_{\mathcal{H}} \right)_{i=1}^{|\mathcal{U}|} \right\|
\end{equation}
\end{lemma}
This lemma is intuitive, given the spectral representation of functions in an RKHS, and the proof is provided in Appendix~\ref{app:proof_of_lemma1}. In general, this lemma implies that in an RKHS with a differentiable kernel function, the approximation error of the functions can bound the gap between their gradients. Since the domain of $\zeta$, i.e., $\Omega$, is a bounded region and both gradients are continuous functions, we can directly deduce that the norm of the kernel gradient, $\|\nabla_{\zeta_i} K\left(\zeta, \cdot\right)\|_{\mathcal{H}}$, is bounded. In fact, for common kernel functions, such as the Gaussian and Laplacian kernels, this gradient remains bounded even for $\zeta \in \mathbb{R}^{|\mathcal{U}|}$.

For simplicity, we assume that $p(u \mid \zeta)$ lies in a common RKHS $\mathcal{H}$ for all $u \in \mathcal{U}$. In fact, $p(u \mid \zeta)$ could lie in different RKHSs for different values of $u$, and our proof still holds as long as our estimate $q(u \mid \zeta)$ lies in the same RKHS as $p(u \mid \zeta)$. The only modification required in this case is taking the supremum over $u \in \mathcal{U}$ in the necessary positions.
\begin{assumption}[Realizability]\label{assumption_realizability}
For all $u \in \mathcal{U}$, the true performative distribution $p(u \mid \zeta)$ and our estimate $q(u \mid \zeta)$ lie in a common RKHS $\mathcal{H}$ with kernel function $K\in C^{2s}(\Omega \times \Omega)$ with $s\geq 2$.
\end{assumption}


\subsection{Convergence guarantee}

Next, we introduce the necessary technical assumptions required for the convergence guarantee. 

\begin{assumption}[Technical assumptions for convergence]\label{technical_assumption}  $\phantom{=}$ 
        
\begin{enumerate}
    \item The kernel gradient is uniformly bounded. $\underset{i}{\max}\left\|\nabla_{\zeta_i} K(\zeta, \cdot)\right\|_{\mathcal{H}} \leq G_K$.
    \item The feature map of RKHS is uniformly bounded. $\|\varphi(\zeta)\|_{\mathcal{H}} \leq R$.
    \item The probability mass function of performative distribution is uniformly lower bounded. There exists $\iota >0 $ s.t. $p(u \mid \zeta)\geq\iota$ and $q(u \mid \zeta) \geq \iota$.
    \item The true performative gradient is uniformly bounded. $\|\nabla_\zeta p(u \mid \zeta)\| \leq G_p$. 
    \item The gradient of policy function is bounded. $\|\nabla_\theta \pi_\theta(x)\| \leq G_\pi$.
    \item The CATE is bounded. $|\tau(x)| \leq B_\tau$.
    \item The estimation error of CATE converges in expectation over the distribution $\mathcal{D}_0$.  $|\mathbb{E}_{X\sim\mathcal{D}_0}[\tau(X) - \hat\tau(X)]| \leq \epsilon_\tau$.   
    \item The performative policy value $V(\pi_\theta)$ is $l$-smooth in $\theta$ and concave in $\theta$.
\end{enumerate}
\end{assumption}
We clarify that most of the boundedness assumptions are applied in a uniform way, meaning they hold for all possible values of $u$, $\zeta$, $\theta$, and $x$. In addition, we only assume the estimation error of CATE converges in expectation, which can be directly derived from classical results~\cite{nie2020cate}. The $\mathcal{D}_0$ represents the data distribution during warm-up stage, where the CATE estimator $\hat\tau$ is trained. We leave the details of subtleties involved here into the Appendix. At last, the norm for functions is $|\cdot|_\mathcal{H}$, while the Frobenius norm is used for vectors and matrices. For consistency, we postpone the discussion of these assumptions to the end of this section.

With these assumptions established, we are now prepared to introduce the convergence guarantee of our strategic policy gradient. Specifically, we consider the time step beginning at the point when the warm-up stage has concluded and our performative gradient estimate is applied.

\begin{theorem}[Convergence of strategic policy gradient]\label{theorem1}
    With learning rate $\eta < \frac{2}{l}$, we have the iterates of true performative gradient satisfying:
    \begin{equation}
    \min_{1\leq t\leq T} \|\nabla_\theta V_t\|^2  \leq \frac{\frac{2}{T}B_\tau + |l\eta^2-\eta|G_VG_E  +\frac{l\eta^2}{2} G_E^2 + 
    \frac{l\eta^2\kappa^2\log (1/\delta_1)}{2n}
    }
{(\eta-\frac{l\eta^2}{2})}
\end{equation}
    holds with probability $1-T\delta_1\delta_2$. Here, $\kappa$ is a constant from concentration inequality. $\epsilon$ is the estimation error $\mathbb{E}[|q(u\mid \zeta)-p(u\mid \zeta)|]$, which satisfies:
    \begin{equation}
        \epsilon = O\left(n^{-1/4}\left(\sqrt{\|p\|_\mathcal{H}/\iota} + \sqrt[4]{\log(1/\delta_2)}\right)\right)
    \end{equation}

    Moreover, 
    \begin{equation}
        G_V = G_\pi B_\tau + B_\tau \frac{G_p G_\pi\sqrt{\mathcal{|U|}}}{\iota}
    \end{equation}
    and 
    \begin{equation}
        G_E = \epsilon_\tau G_\pi\frac{(1+G_p  \sqrt{\mathcal{U}})}{\iota}  + \epsilon B_\tau \left(\frac{G_p}{\iota^2} + \frac{G_K\sqrt{|\mathcal{U}|}}{\iota} \right) G_\pi\sqrt{|\mathcal{U}|}
    \end{equation}
    are upper bounds on $\|\nabla_\theta V(\pi_\theta) \|$ and $\|\mathbb{E}[\hat g] - \nabla_\theta V(\pi_{\theta})\|$, respectively.
\end{theorem}

Theorem~\ref{theorem1} guarantees convergence to a critical point, with the detailed proof provided in Appendix~\ref{app:proof_of_theorem1}. The upper bound approaches $0$ as $\epsilon$ and $\epsilon_\tau$ converge to $0$ and $T$ tends to infinity. Notably, $\epsilon$ and $\epsilon_\tau$ naturally diminish as the sample size per epoch, $n$, increases. Moreover, under a stronger regularity assumption such as $\alpha$-strongly concave on $V(\pi_\theta)$, one can directly bound the distance of $\theta_t$ and performative optimal solution.

Given the complexity of performative learning, several technical assumptions are necessary. We now revisit and discuss some of these assumptions.

To begin, the first two assumptions are imposed on the kernel function. These assumptions are relatively weak and easy to verify. In fact, most common kernel functions, except for the polynomial kernel, typically satisfy these conditions. The upper bound on the norm of the feature map, $R$, is absorbed into the $O$ notation of $\epsilon$, as it is a constant determined by kernel function and not concerned with the asymptotics. Therefore, it does not appear explicitly in our theorem.

Second, since we apply a REINFORCE-type estimator for the policy gradient, as shown in Equation (\ref{eq:full_grad}), a logarithm of probability mass function is introduced. 
Consequently, we must impose a lower bound, which, in our view, cannot be bypassed, given that the gradient used in the algorithm inherently involves this logarithmic term. Moreover, this assumption plays a critical role in theoretically addressing the distribution shift: the lower bound actually implies overlap, which is important for the convergence of function approximation when transitioning from $L_2(\mathcal{D}(\pi_{\theta_1}))$ to $L_2(\mathcal{D}(\pi_{\theta_2}))$.
At last, this assumption can be realized through interesting ways: if, for all $u \in \mathcal{U}$, a small proportion of agents always retain the original feature $u_0 = u$ and resist incentivization—being "always-takers" or "never-takers" in causal inference terminology—then this assumption holds naturally. Additionally, if the principal has absolute control over the information channel and chooses not to inform a randomly selected small proportion of agents about the deployed policy, this assumption remains valid.

Third, we discuss the assumption on concavity of the performative policy value, which is closely related to the convexity of performative risk in the performative prediction literature. On one hand, we assert that assuming convexity or concavity of performative objective function is standard practice to ensure convergence guarantees in performative learning \cite{perdomo2020performative, izzo2021learn}, as well as in our procedures. On the other hand, there are ongoing efforts to identify conditions under which performative risk becomes convex \cite{miller2021outside, cyffers2024optimal}. However, the current results are largely confined to simple cases. In general, performative learning remains a challenging non-convex optimization problem.


\section{Practicality of Discretization}\label{sec:discretization}

\subsection{Discussion on discrete manipulatable features}\label{sec:discrete}

We now elaborate on the validity of the discreteness of $\mathcal{U}$ in Assumption~\ref{assumption1}, which, while initially appearing restrictive, is crucial to our methodology. We argue that discretizing $\mathcal{U}$ is not only a practical approach but also important for making performative learning feasible, without depending on strong parametric assumptions that are often fragile in real-world applications.

First, considering bounded rationality, we argue that agents are unlikely to engage in non-convex continuous optimization involving the policy network $\pi_\theta$. In performative policy learning, the principal interacts with a population of agents—such as a bank interacting with loan applicants or a platform interacting with consumers. This scenario differs significantly from traditional game theory settings, including multi-agent or principal-agent cases, where participants are assumed to engage in strategic interactions with full rationality despite information asymmetry. In practice, the optimization process that most agents realistically undertake can often be reduced to \textbf{selecting the maximum from a finite set of values}.

Second, we turn to the technical rationale. To maintain generality, we begin by considering the continuous case where $u \in [0,1]$. On one hand, in this setting, the input to the behavior model $h_\gamma$ becomes functional data, represented as $\pi_\theta(\cdot, v)$. Both classical statistics and modern deep learning approaches~\cite{li2010uniform,yao2021deep} typically require such functionals to be observed at discrete points for representation.
On the other hand, on the output side, we will encounter a task of continuous density estimation, which is far from tractable in this context. In contrast, when $\mathcal{U}$ is discrete, the observed submitted feature $u^\prime$ enables a more practical approach: a classification task that implicitly estimates the conditional probability mass function.

\textbf{Remark}. When the space of manipulatable features is low-dimensional but includes continuous components, discretization can be performed using techniques such as equal spacing or clustering. In the case of clustering, the resulting labels can be interpreted as agent types, thereby forming the desired discrete space $\mathcal{U}$. However, this approach introduces additional complexity to the policy learning process and may obscure the core contributions of our work. Moreover, it goes beyond the scope of this paper. As such, we leave this as a potential direction for future research and do not address it further in this paper.


\subsection{Incentivize discretization}

In addition to discussing the validity of Assumption~\ref{assumption1}, we demonstrate that, under certain weak regularity assumptions on the agent's micro-level mechanism, the principal can employ a piecewise constant policy with knots $a_1, a_2, \ldots, a_k$ to incentivize agents who manipulate their features to approach these knots.

In the illustration below, we use basic notations from microeconomics, with $\succ$ denoting the agent's (strict) preference regarding the decision to manipulate, and $U$ representing the utility function, which should not be confused with the space of manipulatable features $\mathcal{U}$. For simplicity, we consider the case where $u \in [0,1]$ is the continuous manipulatable feature, while $v$ remains fixed. The agent's original manipulatable feature is still denoted as $u_0$.


Next, we define the utility function as $U=U(u_0,u,v,\pi)$ denotes the utility that an agent with original feature $u_0$ and fixed feature $v$ receives if he decides to move to $u$, given the deployed policy $\pi$. For example, one can still consider the formulation of classic strategic classification as illustrated in Equation (\ref{eq:strategic_classification}), where
\begin{equation}\label{eq:utility}
    U(u_0, u, v,\pi_\theta) = \pi_\theta(u,v) - c\|u-u_0\|_2
\end{equation}

Now we introduce the necessary assumptions:
\begin{assumption}[Regular manipulation]\label{assumption3}
    Given the agent with original feature $u_0$, fixed feature $v$ and deployed policy $\pi$,
    \begin{enumerate}
        \item If there are two alternative point $u_1,u_2$ with $U(u_0,u_1, v,\pi)>U(u_0,u_2, v,\pi)$, then it must hold that $u_1 \succ u_2$.
        \item If there are two alternatives points $u_1,u_2$ with $\pi(u_1) = \pi(u_2)$, and $|u_0-u_1| < |u_0-u_2|$, then it must hold that $U(u_0,u_1,v, \pi)>U(u_0,u_2,v ,\pi)$, and thus $u_1\succ u_2$.
    \end{enumerate}
\end{assumption}

The first part of this assumption states that the agent will strictly prefer the option with higher utility. The second part asserts that, given two alternatives with the same propensity, the agent will strictly prefer the closer one. Note that we do not impose any specific restrictions on how ties are broken.

We now formalize the piecewise constant policy. Given knots $a_1, a_2, \dots, a_k$, we define the intervals $I_j = [a_j, a_{j+1})$ for $j = 1, 2, \dots, (k-1)$, and $I_0 = [0, a_1)$, $I_k = [a_k, 1]$. These intervals form a partition of $[0, 1]$, and their indices represent the discretization into $(k+1)$ levels.

By a piecewise constant policy, we mean that if $u \in I_j$, then we have:
\begin{equation}
    \pi(u,v) = \pi_j(v) 
\end{equation}

Given Assumption~\ref{assumption3}, we can construct the following proposition:
\begin{proposition}\label{prop2}
    Under Assumption \ref{assumption3}, if a piecewise constant policy with knots $a_1, a_2, \dots, a_k$ is deployed, then all agents who choose to manipulate their feature $u$ will move to these knots or to positions that approach the knots infinitely.
\end{proposition}

The proof of this proposition is provided in Appendix~\ref{app:proof_of_prop2}, which is very intuitive. Nevertheless, the underlying insight here is profound: through deploying a policy that is not fully personalized, we can simplify the manipulation patterns and make performative learning feasible with tiny loss of precision. In fact, despite the distortion in the policy release, we can incentivize those responsive agents to move to the predefined knots. 

Nonetheless, we cannot precisely characterize the influence of discretizing $\mathcal{U}$ on the accuracy of the estimated performative gradient. Specifically, there are also agents who may not decide to manipulate if the policy is poorly designed or if the manipulation cost is too high for them. Generally, in the context of repeated Stackelberg games, where either the principal or the agent is learning, characterizing the influence of discretization is quite challenging~\cite{zhu2023sample}. Therefore, we empirically examine the impact of discretization with various granularities in our simulation study.



\section{Experiment with Synthetic Data}
\subsection{Basic setting}

We first carry out the experiments on synthetic data. We set the scale as $|\mathcal{U}|=5, \dim \mathcal{V}=20$. We generate fixed feature $v$ with Gaussian distribution and generate $u$ through $\mathbb{P}(u\mid v) = \operatorname{Softmax}(Wv)$, where $W$ is a predefined transformation matrix. All of relevant parameters are presented in the Appendix \ref{app:params_syn}.

We then generate potential outcomes with linear dependence $u,v$ with a higher positive weight for larger $u$, which means good policy would incentivize the agent to increase his $u$ for better outcomes. Specifically, we choose a classical linear model
\begin{equation}
    Y_i(z_i) = (z_i+1)\langle x_i, \beta \rangle + \epsilon_i
\end{equation}
where $\epsilon_i$ is i.i.d. standard Gaussian noise. 

We generate $v$ from standard multivariate Gaussian distribution, and the discrete feature $u$ is generated by
\begin{equation}
    \mathbb{P}(u\mid v) = \operatorname{Softmax}(Wv)
\end{equation}
Both the weight vector $\beta$ and matrix $W$ mentioned above are predetermined by sampling from Gaussian distribution with fixed random seed, so that kept unchanged throughout the simulation study. 

Moreover, concerning the manipulation mechanism, we mainly consider the classic best-response mechanism:
\begin{equation}\label{eq:mechanism_synthetic}
    u^\prime = \operatorname{argmax}_{u} \pi_\theta(u,v)-c|u-u_0|
\end{equation}
We test three levels of $c \in \{0.05, 0.1, 0.15\}$ and report the results for $c = 0.1$, with the others provided in the appendix. We note that these three levels have minimal impact on the optimal solution when $|\mathcal{U}| = 5$. Instead, the cost coefficient primarily affects the initial phase, where the policy is not yet well-learned, leading to fewer agents choosing to manipulate as the cost coefficient increases. Since the goal of the synthetic experiment is primarily to provide a proof of concept, we defer the more challenging settings with larger $|\mathcal{U}|$ and higher manipulation cost coefficients to the semi-synthetic experiment later.

In addition, we consider another two challenging manipulation mechanisms, beyond the best-response mechanism. The first one is the noisy utility, where the propensities perceived by an agent incur noise, which may stem from measurement errors or limited cognitive abilities.
\begin{equation}\label{eq:noisy_manipulation}
    u^\prime = \operatorname{argmax}_{u} \pi_\theta(u,v) + \epsilon_\pi-c|u-u_0|
\end{equation}
here $\epsilon_\pi \sim \mathcal{N}(0, c^2)$.

The second mechanism is the softmax utility, representing a random manipulation pattern. This corresponds to a novel manipulation pattern, where agents would still change their $u$ to different levels even given the same $v$ and $\pi_\theta$. 
\begin{equation}\label{eq:softmax_manipulation}
    u^\prime \sim \operatorname{Softmax}\left(5 (\pi_\theta(u,v) -c|u-u_0|)\right)
\end{equation}
We use a multiplier of $5$ to address the scale issue; otherwise, the utilities would likely be numerically similar, causing $u^\prime$ to follow an approximately uniform distribution.

For comparison, we evaluate the proposed approach against several baseline methods. Additional details about these baselines are provided in Appendix \ref{app:baseline}.
\begin{enumerate}
    \item \textit{Cutoff Rule}, which is a cutoff rule and optimal without strategic behavior. Formally, the propensity only depends on the sign of CATE: $\pi(x)=\mathbb{I}(\tau(x)>0)$. 
    \item \textit{Vanilla Policy Gradient}, which ignores the impact of policy on data distribution, and updates the model parameters with gradient $\mathbb{E}_{x \sim p\left(\cdot\right)}\left[\nabla_\theta \pi_\theta(x) \tau(x)\right]$. 
    \item \textit{End2end Policy Gradient}, which models $p\left(u \mid v, \pi_\theta\right)$ with $p\left(u \mid v, \theta\right)$. We construct this baseline method to demonstrate the benefit of taking $\pi_\theta$ as intervention instead of its parameter $\theta$.
\end{enumerate}
We note that the second baseline, \textit{Vanilla Policy Gradient}, is adapted from the repeated gradient descent approach \cite{perdomo2020performative}, with the key difference being that we apply gradient ascent to maximize the policy value in this context. The third baseline, \textit{End2end Policy Gradient}, is termed "end-to-end" because it incorporates the concept of the distribution map $\mathcal{D}(\theta)$ and directly models the performative distribution $p(u \mid v, \pi_\theta)$ as $p(u \mid v, \theta)$. For comparison, the basic settings for all three gradient-based methods remain nearly identical, with the primary distinction being the gradient used after the warm-up stage (the 30th epoch). Finally, the CATE in the three gradient-based methods is estimated, while the cutoff rule uses the oracle CATE.

\subsection{Result and discussion}\label{sec:4.2}

\begin{figure}[t]
\centering
\begin{subfigure}{0.45\columnwidth}
    \centering
    \includegraphics[width=\columnwidth]{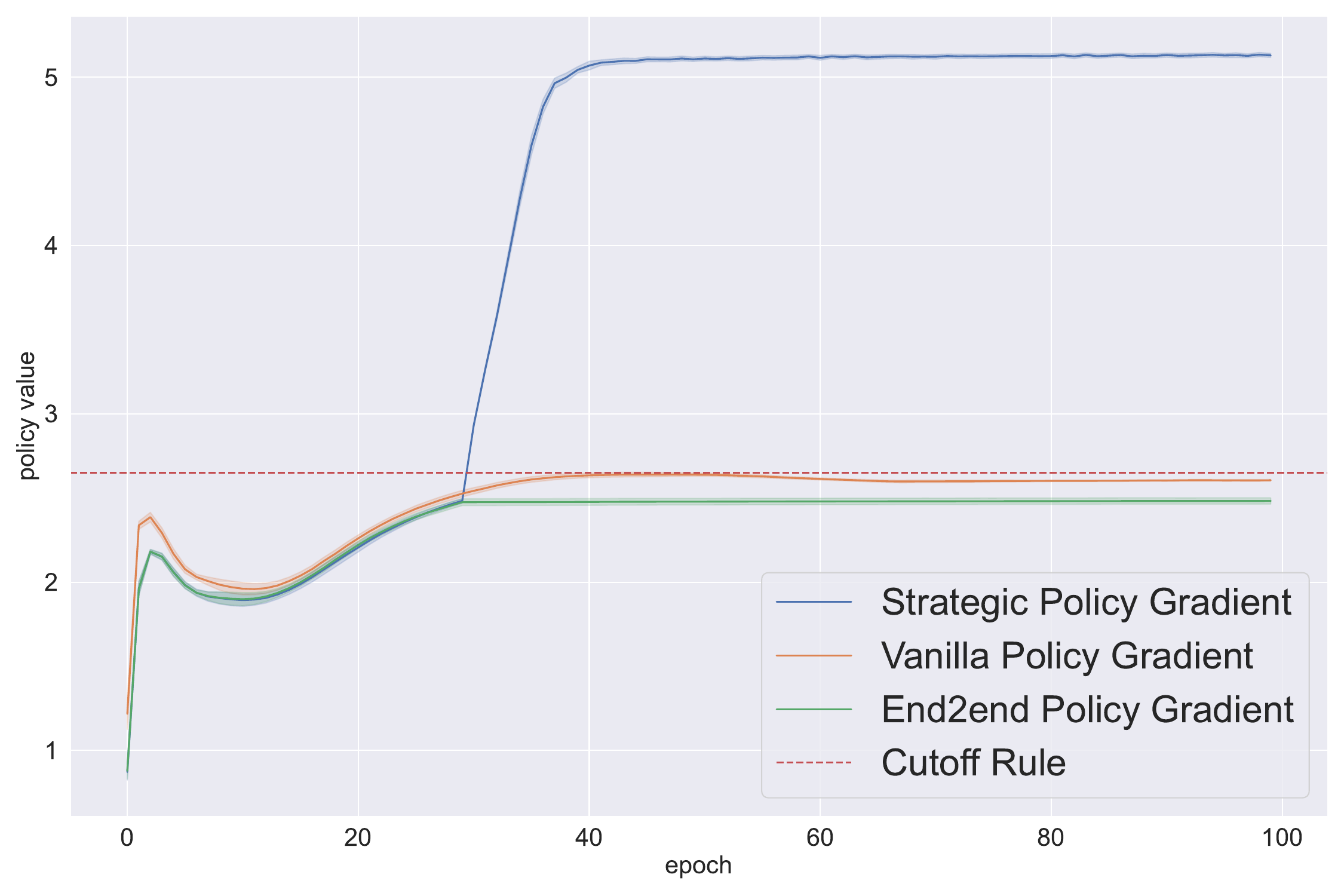}
    \caption{}
    \label{fig:syn_subfig1}
\end{subfigure}
\hfill
\begin{subfigure}{0.52\columnwidth}
    \centering
    \includegraphics[width=\columnwidth]{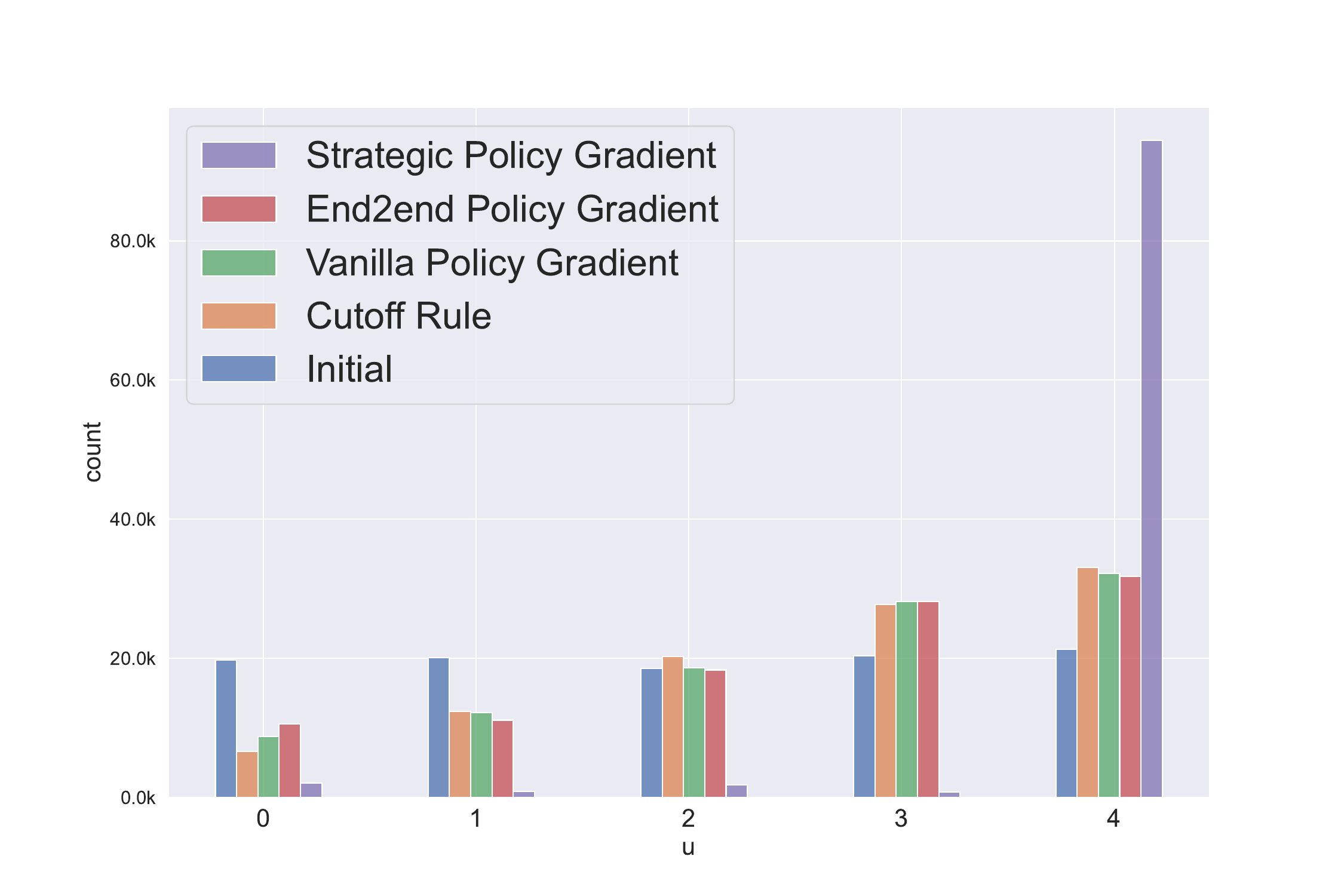}
    \caption{}
    \label{fig:syn_subfig2}
\end{subfigure}
\caption{The left figure presents the curve of policy value in \textbf{synthetic} experiment with $c=0.1$, and right figure presents the original distribution of $u$ and that after manipulation induced by policies.}
\label{fig:policy_curve_synthetic}
\end{figure}

The policy value curves with $c=0.1$ are shown in Figure~\ref{fig:syn_subfig1}. The distributions of the manipulatable feature, both before and after incentivization by each policy, are displayed in Figure~\ref{fig:syn_subfig2}. Detailed results can be found in Appendix \ref{app:result}. The error bars represent the $\pm$ standard error. All methods were evaluated using 10 successive random seeds.

In Table~\ref{table:semi_synthetic_main_paper}, we report the policy value at both the best point in the entire trajectory and the final epoch. Additionally, we present the proportion of agents who change their $u$ in the column \%CHANGE, as well as the mean signed change in $u$ induced by different policies in the column MOVE, where the average is calculated across all agents.

\begin{table}[t]
\caption{Experiment result on synthetic data with \textbf{$c=0.1$}}
\label{table:synthetic_main_paper}
\begin{center}
\begin{small}
\begin{sc}
\begin{tabular}{lcccc}
\toprule
Method & Policy value (Best) & Policy value (Final epoch) & \%Change & Move \\
\midrule
Cutoff & $2.92 \pm 0.00$ & $2.92 \pm 0.00$ & 37.50 & 0.74\\
Vanilla& $2.96 \pm 0.02$ & $2.89 \pm 0.01$ & 30.82& 0.59\\
End2end & $2.76 \pm 0.06$ & $2.76 \pm 0.06$ &24.29 & 0.48 \\
Strategic& $5.26 \pm 0.03$ & $5.23 \pm 0.03$ & 74.38 & 1.87\\
\bottomrule
\end{tabular}
\end{sc}
\end{small}
\end{center}
\vskip -0.1in
\end{table}


First, our proposed method stands out due to its exceptional performance, maintaining stability in convergence. We also observe that the cutoff policy is a performatively stable solution, as it acts as a fixed point of \textit{Vanilla Policy Gradient}. Our experiments with both synthetic and semi-synthetic data provide empirical evidence of the significant gap between performatively stable and optimal solutions.

We also observe that the performance of \textit{End2end Policy Gradient} is similar to that of \textit{Vanilla Policy Gradient}, which we attribute to the failure in learning the behavior model $h_\gamma$. It is worth noting that the optimizer is reset to apply the specified performative gradient for both \textit{End2end Policy Gradient} and our method, \textit{Strategic Policy Gradient}, while all three gradient-based methods initially use the Adam optimizer~\cite{kingma2017adammethodstochasticoptimization}. The policy value of \textit{End2end Policy Gradient} remains unchanged because the gradient $\nabla_\theta p(u \mid v,\theta)$ is almost zero, whereas the policy value of \textit{Vanilla Policy Gradient} continues to evolve due to the first-order momentum.

In addition, Figure~\ref{fig:syn_subfig2} presents the distribution of the manipulable feature $u$ before and after manipulation, induced by the four methods, respectively. These distributions offer an alternative perspective on the performance of the methods, as an increase in $u$ corresponds to a higher CATE. In line with this, Table~\ref{table:synthetic_main_paper} shows that our strategic method incentivizes the largest proportion of agents to change their feature, with the highest mean movement in $u$.

However, we note that the distribution of post-manipulation features provides only an indirect reference for understanding the method's performance. For example, as described in Equation (\ref{eq:strategic_classification}), propensity vectors $(0, 0.1, 0.2, 0.3, 0.4)$ and $(0.5, 0.6, 0.7, 0.8, 0.9)$ have the same effect on incentivizing an agent's improvement, yet the corresponding policy values can differ significantly.


\subsection{Result with softmax manipulation and noisy utility}

Next, we present results with two additional manipulation mechanisms. This experiment demonstrates that our methodology can accommodate flexible micro-mechanisms by bypassing the need for parametric modeling of such decision processes.

\begin{figure}[ht]
\centering
\begin{subfigure}{0.48\columnwidth}
    \centering
    \includegraphics[width=\columnwidth]{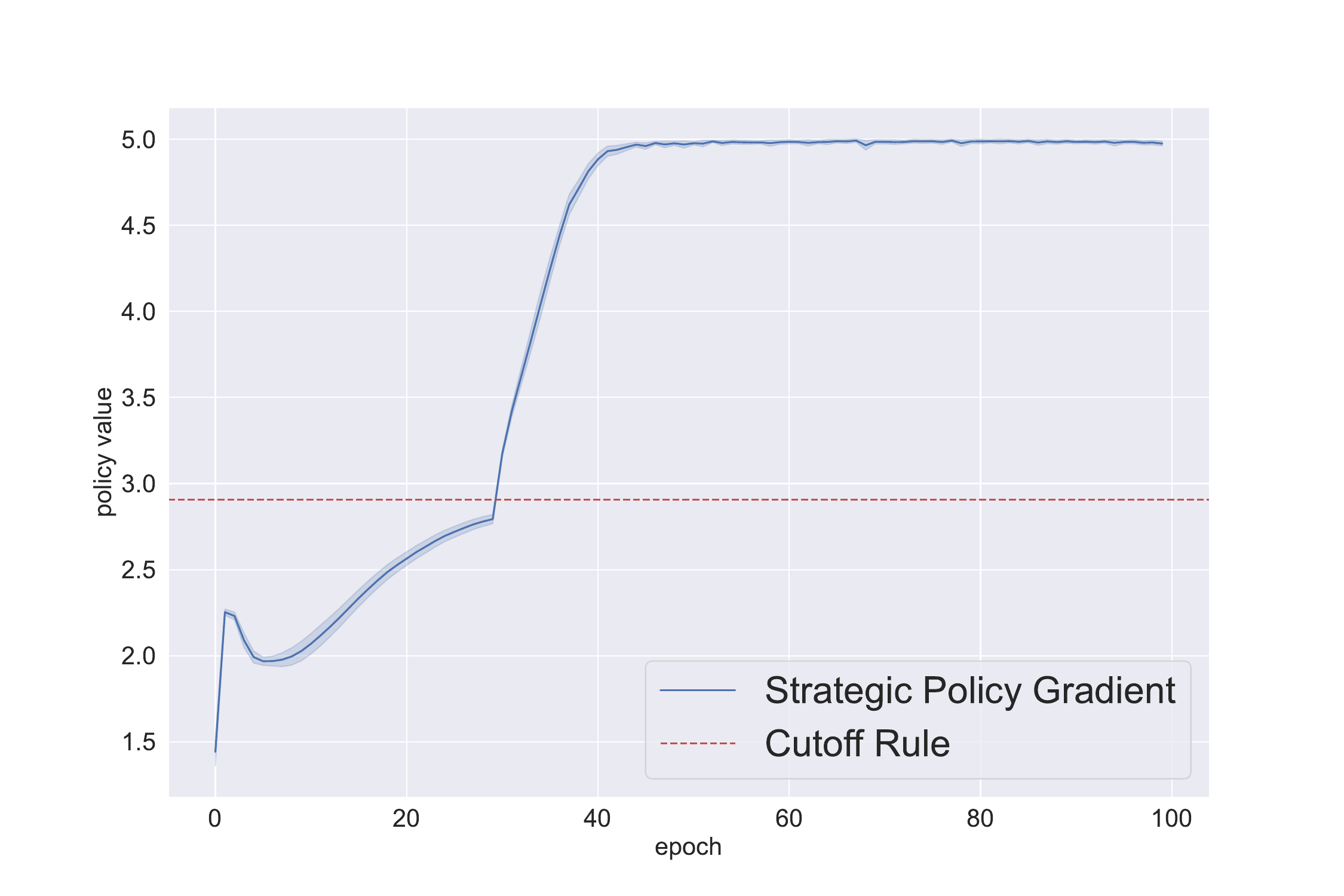}
    \caption{}
\end{subfigure}
\hfill
\begin{subfigure}{0.48\columnwidth}
    \centering
    \includegraphics[width=\columnwidth]{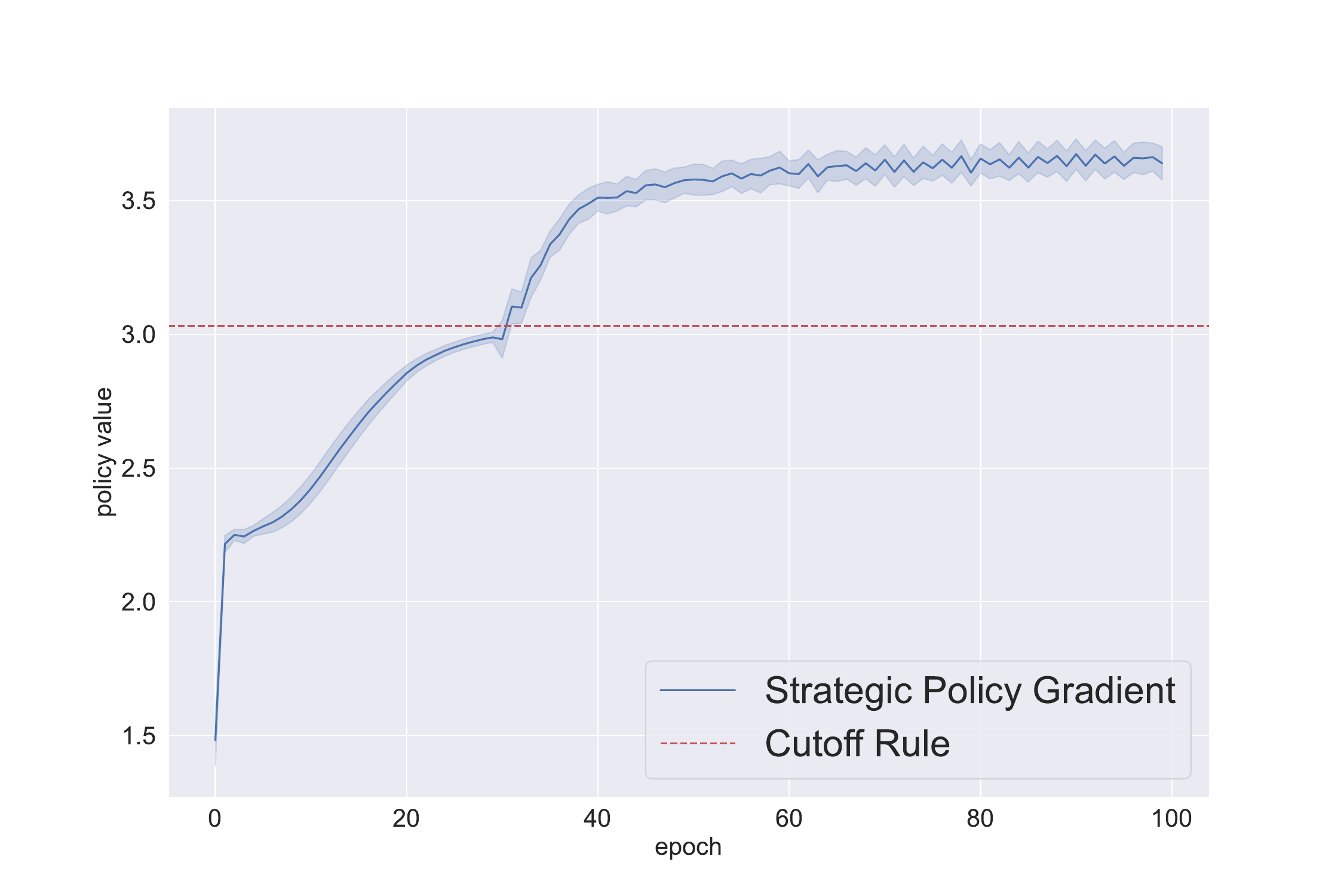}
    \caption{}
\end{subfigure}
\caption{The left figure is the curve of policy value with noisy utility, and the right figure is that with softmax manipulation mechanism.}
\label{fig:softmax_noisy}
\end{figure}

In Figure \ref{fig:softmax_noisy}, we show the performance of the proposed method and the cutoff policy in two more challenging settings: random (softmax) manipulation and noisy utility, as formulated in Equations (\ref{eq:noisy_manipulation}) and (\ref{eq:softmax_manipulation}), respectively. Our results indicate that the strategic policy gradient continues to perform well under these conditions.

\subsection{Gaussian process classifier as behavior model}


In Section~\ref{sec:3.2}, we propose using a neural network, such as a multi-layer perceptron (MLP), as a classifier to learn the performative distribution $p(u \mid \zeta(v, \pi_\theta))$. Since our theoretical analysis relies on function approximation within an RKHS, and MLPs with finite many hidden neurons are generally not contained within an RKHS, we include this subsection for completeness.

We consider the Gaussian process classifier to serve as the behavior model, as Gaussian process can be viewed as an RKHS with a Gaussian measure on top of it. We utilize GPyTorch~\cite{gardner2018gpytorch} to implement Gaussian process classifier with variational approximation. 

\begin{figure}[ht]
\centering
\begin{subfigure}{0.48\columnwidth}
    \centering
    \includegraphics[width=\columnwidth]{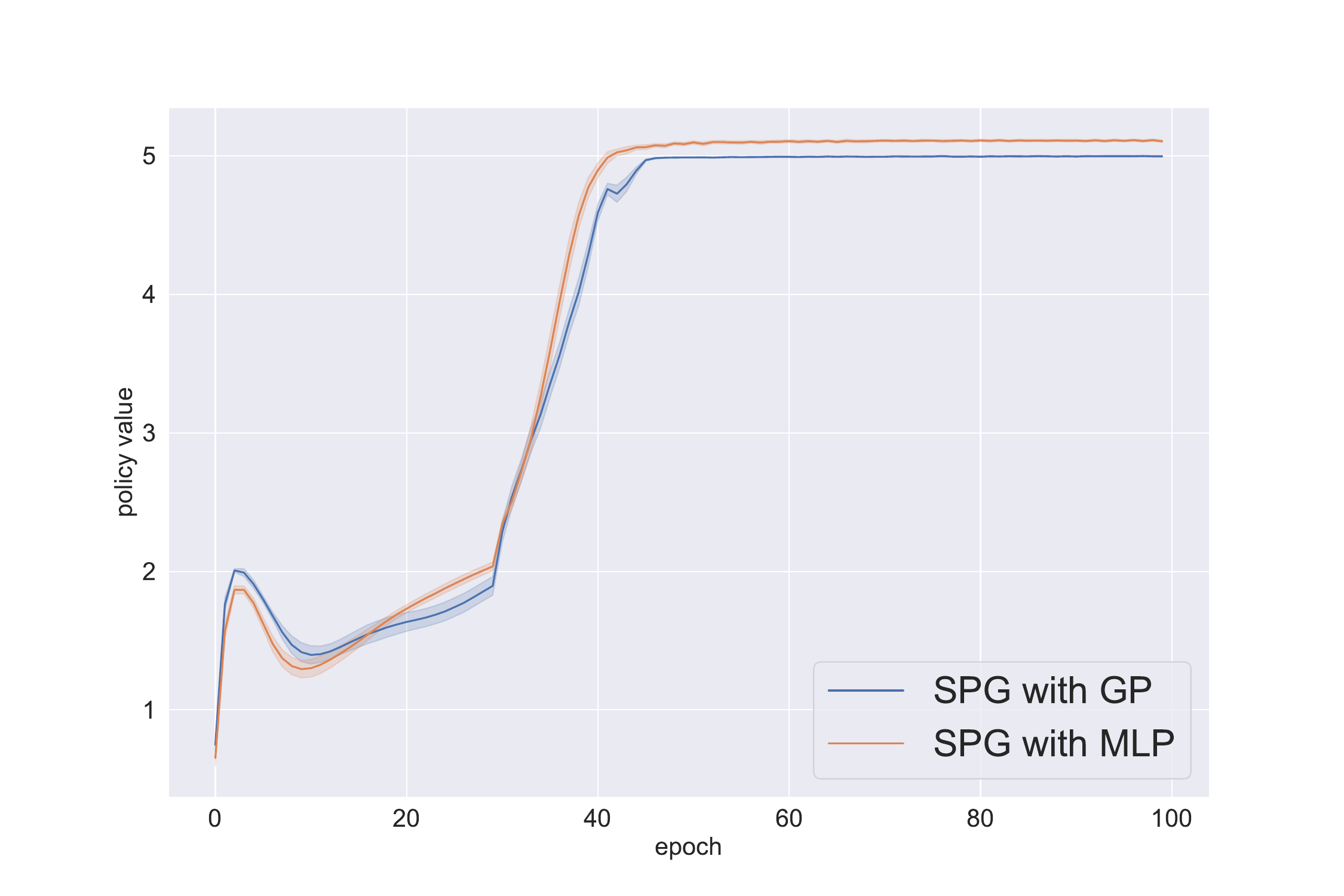}
    \caption{}
\end{subfigure}
\hfill
\begin{subfigure}{0.48\columnwidth}
    \centering
    \includegraphics[width=\columnwidth]{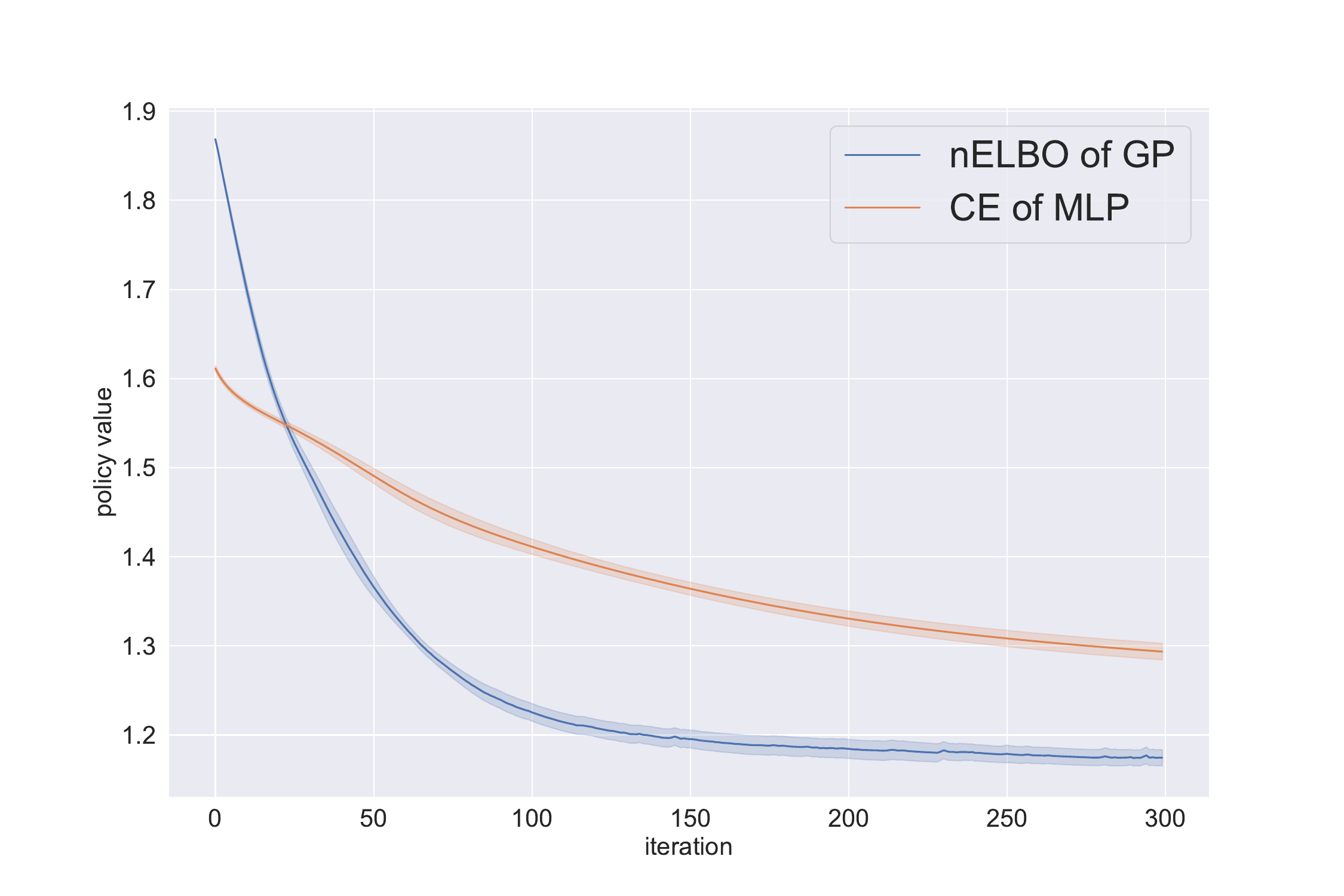}
    \caption{}
\end{subfigure}
\caption{The left figure presents policy value curves of the proposed strategic policy gradient with MLP and Gaussian process classifier as the behavior model. The right figure illustrates the curve of the loss functions the behavior model, i.e. cross-entropy and negative ELBO, respectively.}\label{fig:gp}
\end{figure}

We present the policy value curves and training loss curves with the default best response manipulation and cost coefficient $c=0.1$, in Figure~\ref{fig:gp}. We mention that the loss functions are cross-entropy for MLP and \textit{negative} variational evidence lower bound (ELBO) for Gaussian process classifier. Notably, the Gaussian process classifier performs almost the same as MLP, and rapid convergence is achieved following the warm-up stage. Moreover, we observe that the training of these two models also converges quickly, which indicates that both of them are qualified for being behavior model.

Nonetheless, the Gaussian process classifier is limited by the curse of dimensionality and does not scale well with increasing data dimensions. Since we aim to examine the baseline \textit{End2end Policy Gradient} in the synthetic experiment, where the input dimension is very high (i.e., $\dim(\mathcal{V}) + \dim(\theta)$), we utilize MLP as the behavior model in all other parts of the analysis.

The takeaway message of this subsection is that the behavior model can be chosen flexibly, as long as it possesses sufficient expressiveness to estimate the performative distribution, and supports back propagation with regard to its input, $\zeta$.

\subsection{Examining coarse discretization}\label{sec:coarse_experiment}

In this subsection, we discuss the impact of discretizing the manipulatable feature $u$ at different levels of granularity. On one hand, all measured data can practically be viewed as discrete with a certain level of precision. On the other hand, considering $u$ as continuous complicates both the calculation and validation of the best response. For a manipulation mechanism similar to that in Equation (\ref{eq:mechanism_synthetic}), finding the best response involves solving a highly non-convex optimization problem with no guarantee of obtaining the true optimal solution. Furthermore, these optimization problems must be addressed separately for different agents, as their fixed features vary. Therefore, leveraging continuous $u$ in the simulation study is not only unnecessary but also infeasible.

\begin{figure}[ht]
\centering
\begin{subfigure}{0.48\columnwidth}
    \centering
    \includegraphics[width=\columnwidth]{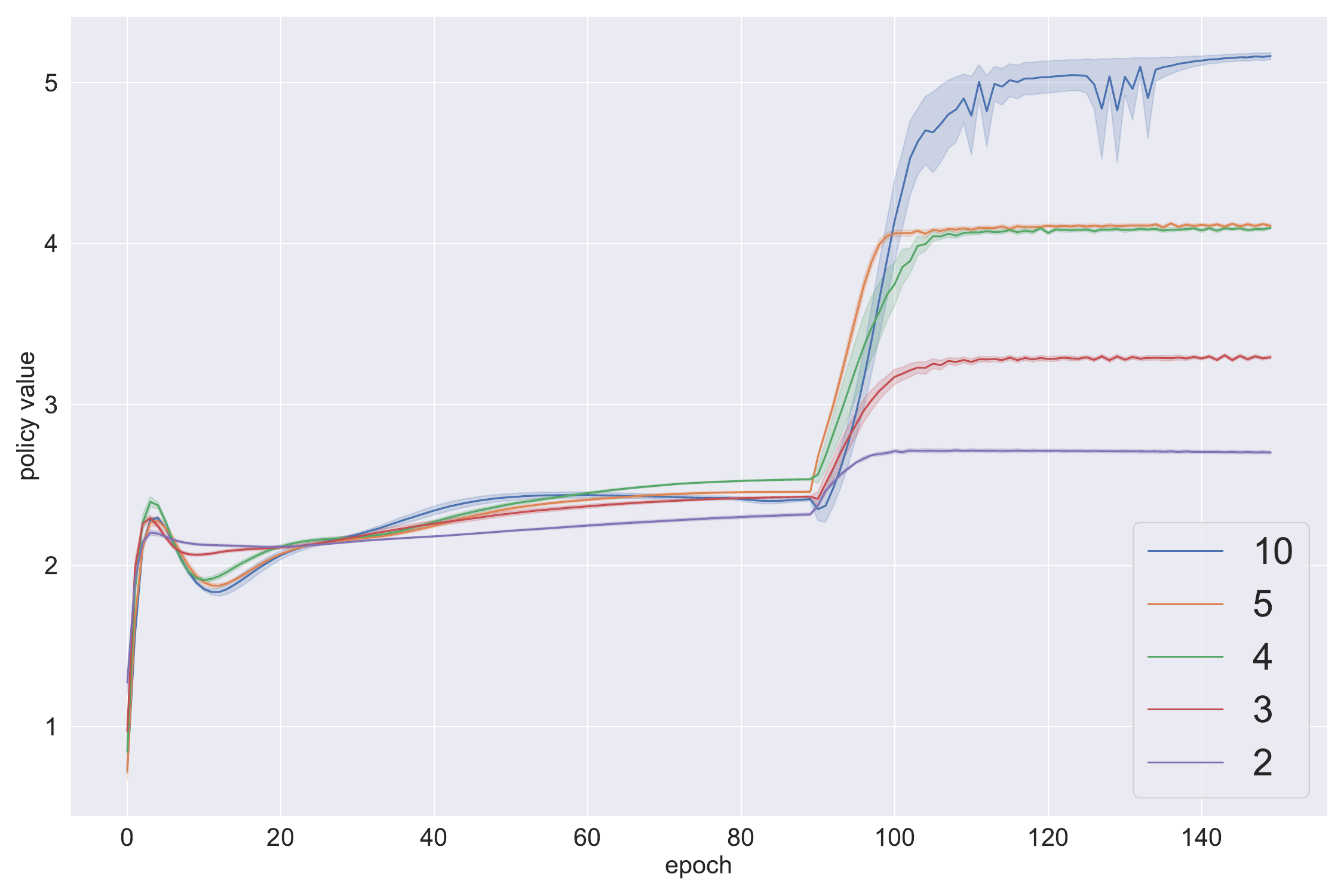}
    \caption{}\label{fig:coarse_discretization1}
\end{subfigure}
\hfill
\begin{subfigure}{0.48\columnwidth}
    \centering
    \includegraphics[width=\columnwidth]{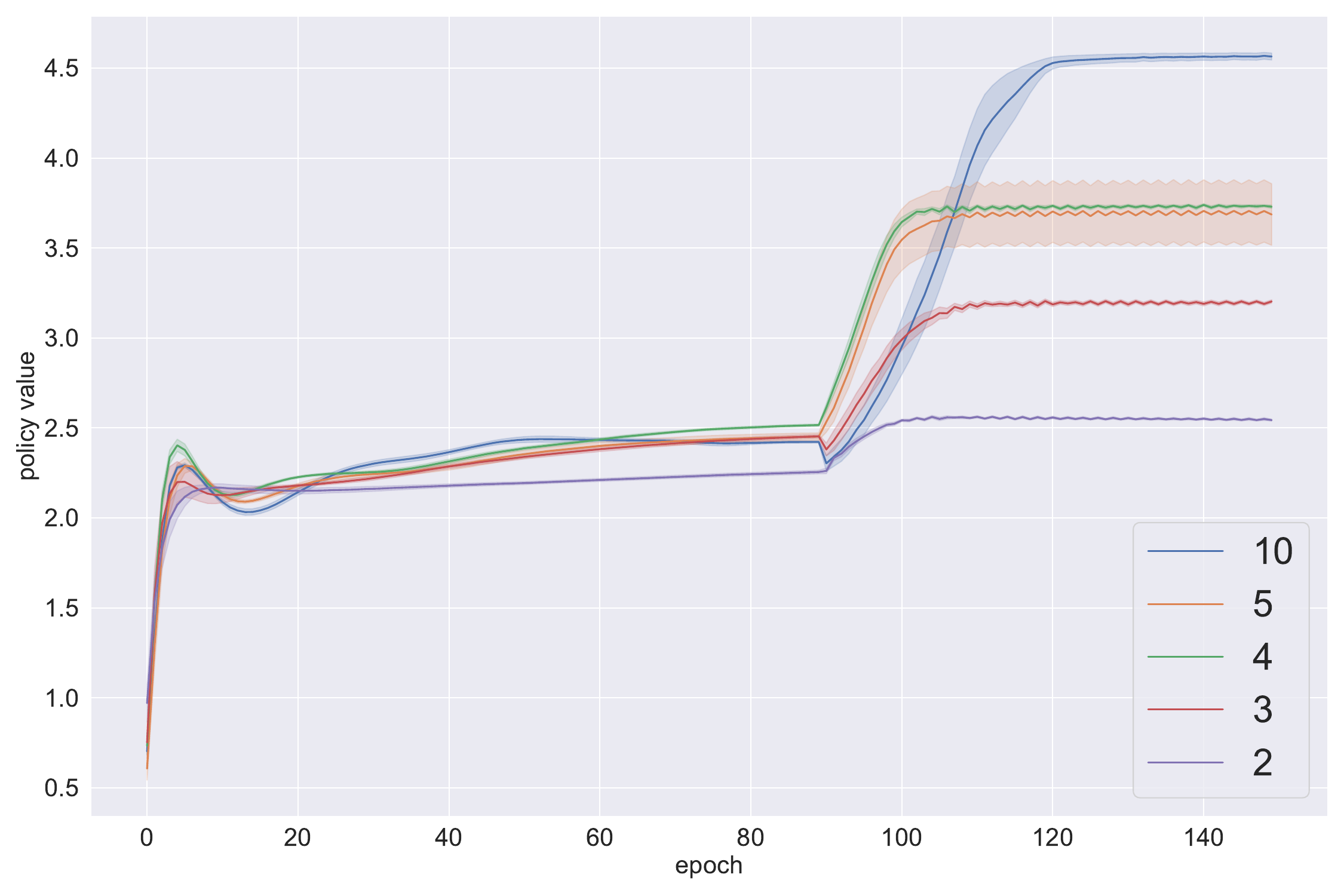}
    \caption{}\label{fig:coarse_discretization2}
\end{subfigure}
\caption{The left figure presents the curves of policy value with the true $|\mathcal{U}|$ being 15 levels, and the right figure presents that with 50 levels. The name of each curve indicates the actually used $|\mathcal{U}|$.}
\end{figure}

Nonetheless, we can set the true cardinality of the manipulatable feature $u$ as $M$ and choose a coarser discretization that further partitions the set $\{1, \ldots, M\}$ into the practically used $\mathcal{U}$, maintaining equal spacing. The practical significance of this procedure lies in the fact that the sample complexity required to accurately estimate the probability mass function can be prohibitively high when $M$ is large. We implement such procedure with $M=15$ in Figure~\ref{fig:coarse_discretization1} and $M=50$ in Figure~\ref{fig:coarse_discretization2}. In each scenario, we present five cases with $|\mathcal{U}|=2,3,4,5,10$, respectively. The mechanism is that agents make decisions according to the true $u$ with $M$ levels, while the principal views its cardinality as $2,3,4,5,10$ levels, and implements policy learning and CATE estimation based on this coarser granularity. 

We mention that since the case of $|\mathcal{U}|=10$ demands more data to train the behavior model, we extend the whole horizon for presentation, with the duration of the warm-up stage being 90 here. Nonetheless, the total number of model deployments needed for convergence after the warm-up stage is still about 30 epochs.

We first observe that when the true cardinality $M=15$, the case of $|\mathcal{U}|=10$ almost maintains the best policy value that is achieved in the case of $M=|\mathcal{U}|=5$, of which the performance is presented in the former subsections. When $M$ increases to $50$, the achieved policy value only decreases by about $0.5$. Hence, we can conclude that the coarser discretization brings little impact on both convergence and value of optimized policy. A similar situation occurs for the cases that $|\mathcal{U}|=3,4,5$, namely, the performance mainly depends on the used $|\mathcal{U}|$ instead of the number of true levels, $M$. In summary, we find that $|\mathcal{U}|=10$ is sufficient in this scenario, even if the number of true discrete levels $M$ is larger than 10.

\section{Experiment with Semi-synthetic Data}
\subsection{Basic setting}\label{sec:5.1}
In this section, we conduct semi-synthetic experiments on a loan dataset.
This dataset is composed of 307,508 loan records, including rich information about the terms of the loans (e.g., loan amount, rate of interest, upfront charge, etc.) and the applicants (e.g., age, status of employment, count of children, etc.). There is a binary label indicating whether the applicant finally defaults. We use this dataset to provide an organic fixed feature $v$ and construct the manipulatable feature $u$ as an external credit score that relies on $v$. We then construct outcome $Y$ as the expected revenue, which corresponds to a much more complex CATE. We split the tabular data into 300 batches that appear in sequence to mimic the sequential experiment setting. We provide all the parameters in Appendix \ref{app:semi-syns}.

In the gross, we discretize the $u$ into $10$ levels, with $v$ and $\tau(x)$ in much more complex forms in this scenario. Thus we use a more complex policy network and omit the \textit{end2end} method since its bad scalability for high-dimensional model parameter $\theta$ and performance shown in synthetic experiment.

\subsection{Data and preprocessing}

The dataset\footnote{The original dataset can be accessed at \url{https://www.kaggle.com/datasets/gauravduttakiit/loan-defaulter}.} we use in this experiment can be accessed in supplementary materials, which has been preprocessed. We adopt the \textit{application\_data.csv} and keep 56 features that are not missing severely. Though some flag features take binary value, the space $\mathcal{V}$ is much more complex than Gaussian distribution in synthetic experiment.

To further combine with our setting, we utilize all of these organic features as fixed feature $v$, and we utilize two Random Forest~\cite{breiman2001randomforest} models with different model complexity to fit the binary label, for generating organic potential outcomes. These two models achieve average precision of 0.99 and 0.9, respectively, and we call them strong and ordinary models for illustration. We treat the probability of no default predicted by the ordinary model as the evaluation score from an external organization that would be taken into consideration by the principal (i.e., the bank). We use this score as the manipulatable feature $u$, and the initial score is deemed as the original manipulatable feature $u_0$. We interpret the manipulation as management of liquidity assets, which is supervised and testified by an external organization. An agent can choose to sell the high-stake asset such as stock to increase the score, which corresponds to an improvement in the user's ability to repay loans and is exactly what the bank wants. The argument above motivates us to model the true probability of default before manipulation, which is 
\begin{equation}
    g(u,v)=\lambda f_{\text{strong}}(v) + (1-\lambda)u=\lambda f_{\text{strong}}(v) + (1-\lambda)f_{\text{ordinary}}(v)
\end{equation}
Here $\lambda\in[0,1]$ is the weight for two evaluations, and smaller $\lambda$ is associated with stronger strategic behavior, thus, we take $\lambda = 0.4$ in this part. After manipulation from $u$ to $u^\prime$, this probability changes into $g(u^\prime,v)=\lambda f_{\text{strong}}(v) + (1-\lambda)u^\prime$. We discretize the domain of $u$, i.e., $\mathcal{U}=[0,1]$ into 10 subintervals, corresponding to $u=0,1,\dots,9$. We also specify the manipulation mechanism as 
\begin{equation}\label{eq:best_response_semi}
    u^\prime = \operatorname{argmax}_{u} \pi_\theta(u,v)-c|u-u_0|
\end{equation}
In this setting, we take the decision of whether to approve a loan to the applicant as the treatment $z_i$, for which $z_i=1$ indicates approval and $z_i=0$ indicates rejection. The principal aims to maximize the expected revenue through a treatment allocation policy. Without loss of generality, we set $Y_i(0)=0$, and the expected revenue for allocating treatment to the agent $i$ is
\begin{equation}\label{eq:outcome}
\begin{aligned}
    Y_i(1) = \left(g(u^\prime_i,v) - \beta(1-g(u^\prime_i,v))\right) \times \text{amt}_i
\end{aligned}
\end{equation}
The $\text{amt}_i$ here is the loan amount of the application. We only keep this feature in the expression of outcome for heterogeneity among agents, with other related features such as the interest rate and term ignored for simplification. We then introduce a multiplier $\beta$ that indicates the loss of loan ending up with default is much more severe contrasting the gain of a normal case. In this part, we will take $\beta=4$, and this choice corresponds to a policy value $-30,570$ for all treatments versus $0$ for all controls.

For preprocessing, we perform label encoding to transform some categorical features into numerical ones. Naturally, there are many more normal cases compared to the default case, so we adopt the synthetic minority oversampling technique (SMOTE)\cite{chawla2002smote} to solve the issue of label unbalance and guarantee the fitted probability to be an effective function of input feature $v$. We also normalize features with mean and standard deviation. 

\begin{figure}[h]
\centering
\begin{subfigure}{0.48\columnwidth}
    \centering
    \includegraphics[width=\columnwidth]{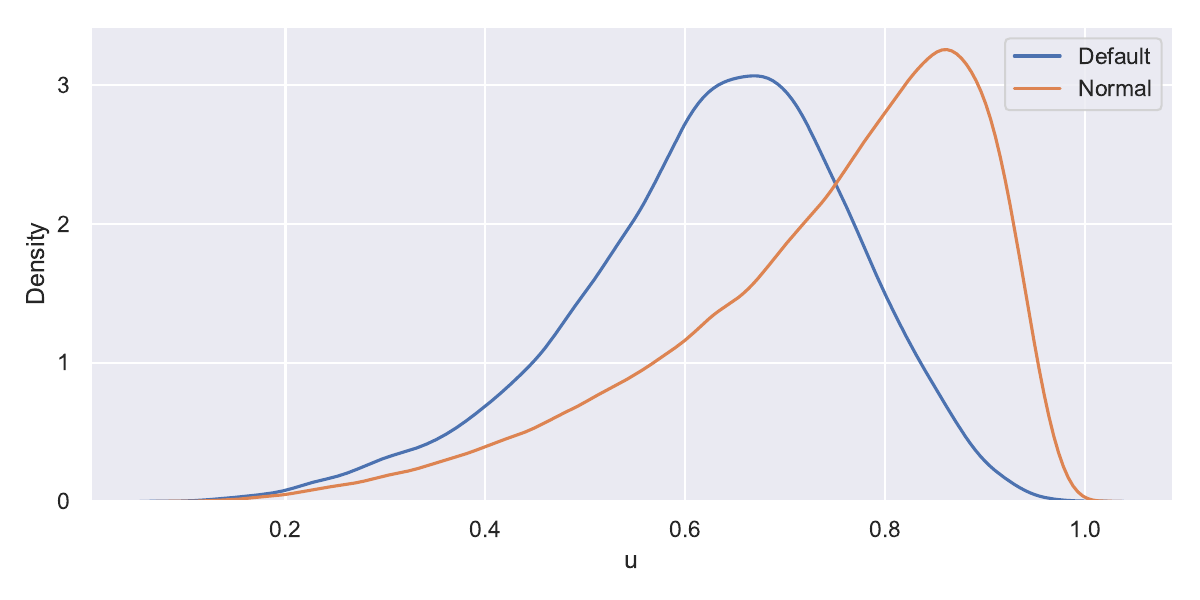}b
    \caption{}
\end{subfigure}
\hfill
\begin{subfigure}{0.48\columnwidth}
    \centering 
    \includegraphics[width=\columnwidth]{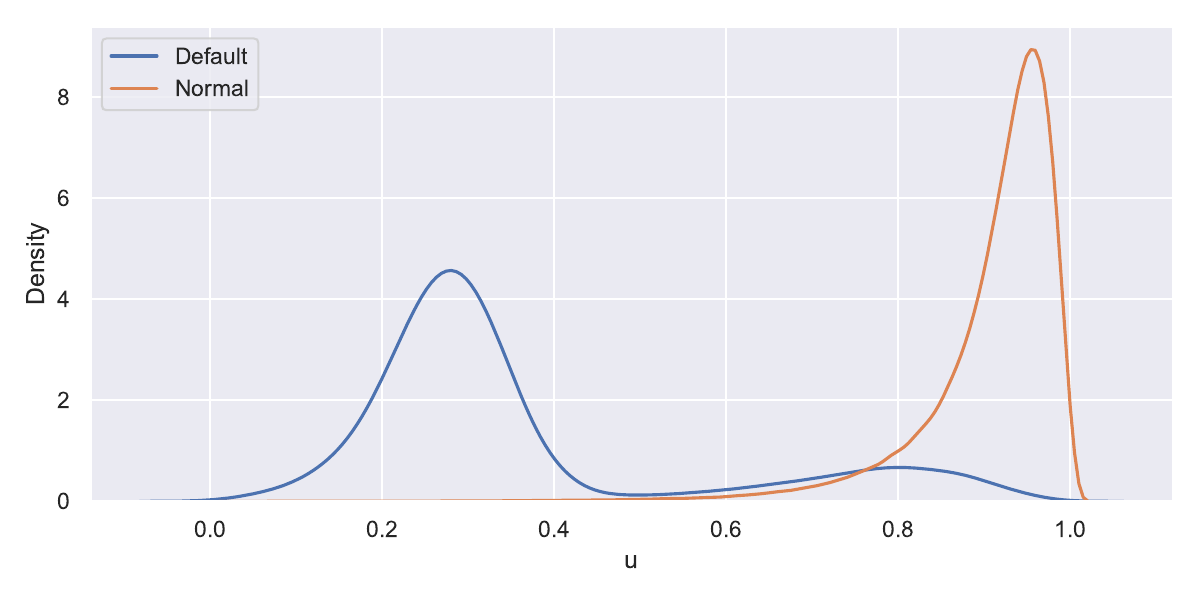}
    \caption{}
\end{subfigure}
\caption{Distribution of predicted probability of no default generated by ordinary (left) and complex (right) random forest models.}
\label{fig:rf}
\end{figure}

For the predictive model, we select the strong model as an ensemble of $200$ decision trees and the ordinary model as an ensemble of $40$ decision trees with a restriction on maximum depth. These two models achieve $0.99$ and $0.9$ F1-score on the test set respectively, with the distribution predicted probability of two classes shown in Figure \ref{fig:rf}. The idea here is that we want to generate the potential outcome with a more reliable default probability, while the original external credit score should be a bit coarser.



\subsection{Features and causal parameters}

\begin{enumerate}
    \item  Manipulatable feature $u$: We discretize the predicted probability of no default generated by the ordinary random forest and let this discretized variable be the manipulatable feature $u \in\{0,1,\ldots,9\}$:
    \begin{align*}
        u=\lfloor 10\times \text{probability of no default} \rfloor
    \end{align*}
    \item Fixed features $v$: In addition to the manipulatable feature $u$, the remaining 55 variables in the dataset are treated as fixed features $v$.
    \item Treatment $z$: The binary variable $z\in\{0,1\}$ represents whether the principal approves the agent's loan request, where $z=1$ indicates approval for granting the loan, and conversely, $z=0$ signifies denial of the loan.
\end{enumerate}

\subsection{Result and discussion}

\begin{figure}[h]
\centering
\begin{subfigure}{0.45\columnwidth}
    \centering
    \includegraphics[width=\columnwidth]{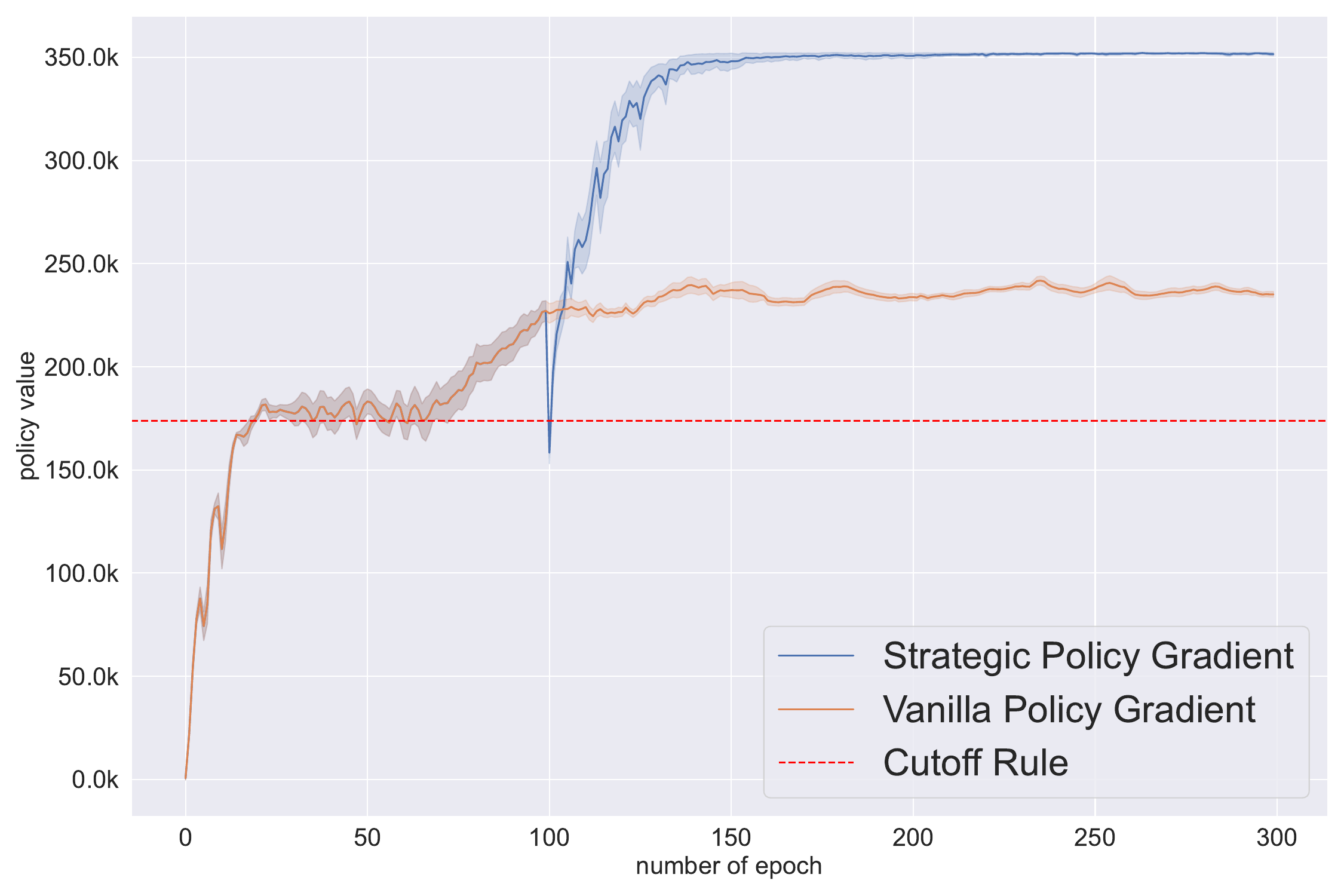}
    \caption{}
    \label{fig:semi_performance}
\end{subfigure}
\hfill
\begin{subfigure}{0.52\columnwidth}
    \centering
    \includegraphics[width=\columnwidth]{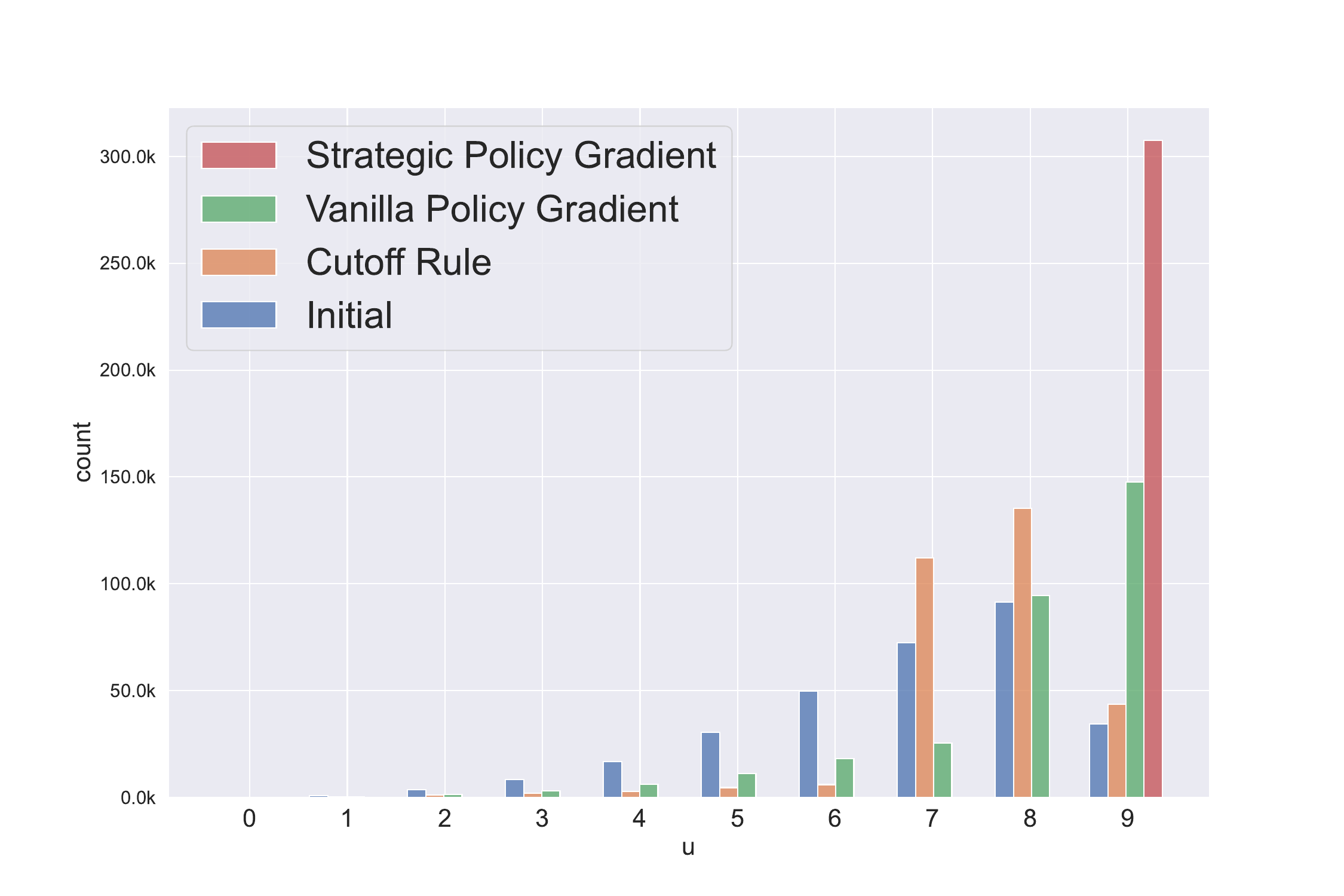}
    \caption{}
    \label{fig:semi_u_distribution}
\end{subfigure}
\caption{The left figure presents the curve of policy value in semi-synthetic experiment with cost coefficient $c=0.1$, and the right figure presents the original distribution of $u$ and that after manipulation induced by policies.}
\label{fig:semi_synthetic}
\end{figure}

\begin{figure}[h]
\centering
\begin{subfigure}{0.45\columnwidth}
    \centering
    \includegraphics[width=\columnwidth]{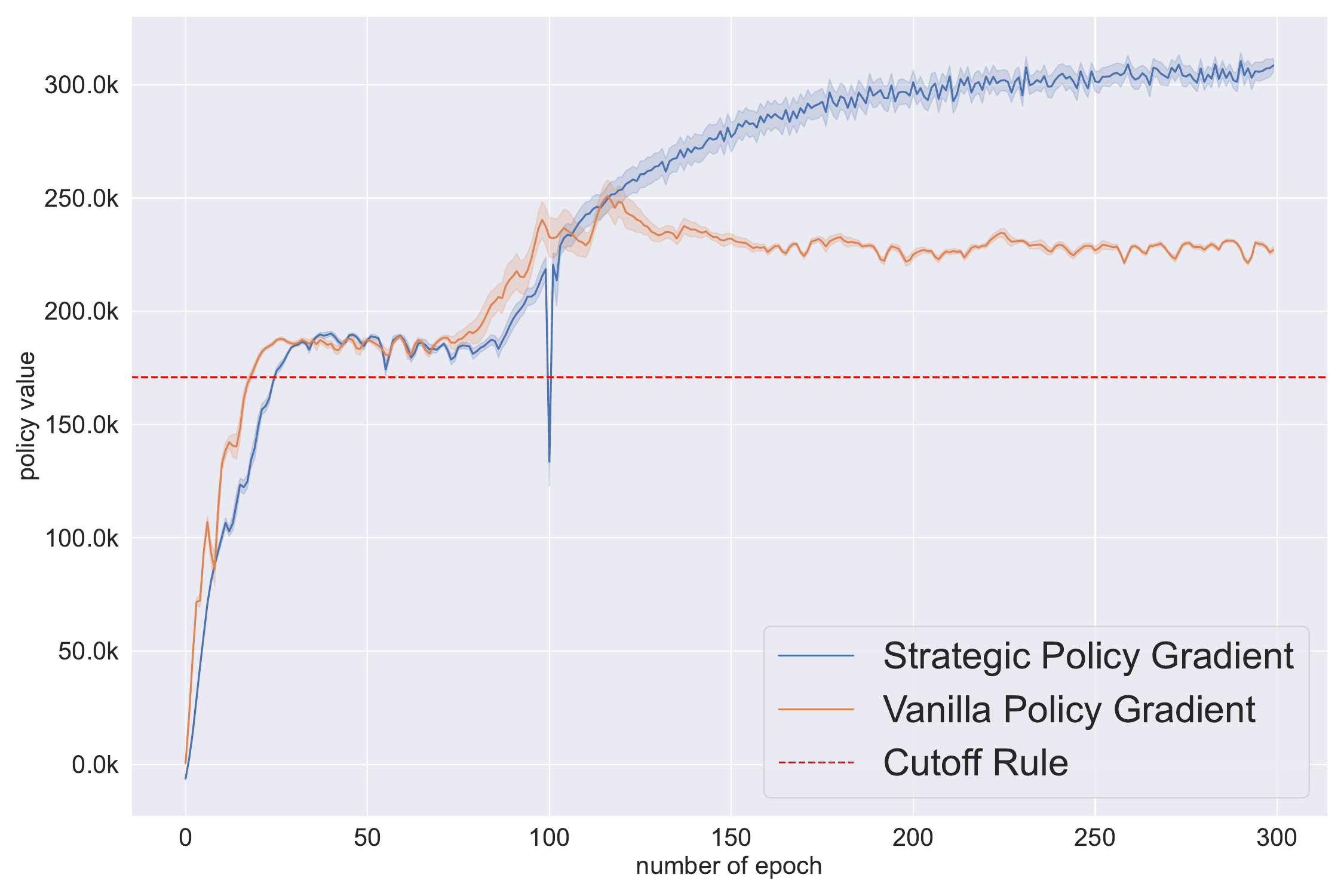}
    \caption{}
    \label{fig:semi_performance_c0.2a}
\end{subfigure}
\hfill
\begin{subfigure}{0.52\columnwidth}
    \centering
    \includegraphics[width=\columnwidth]{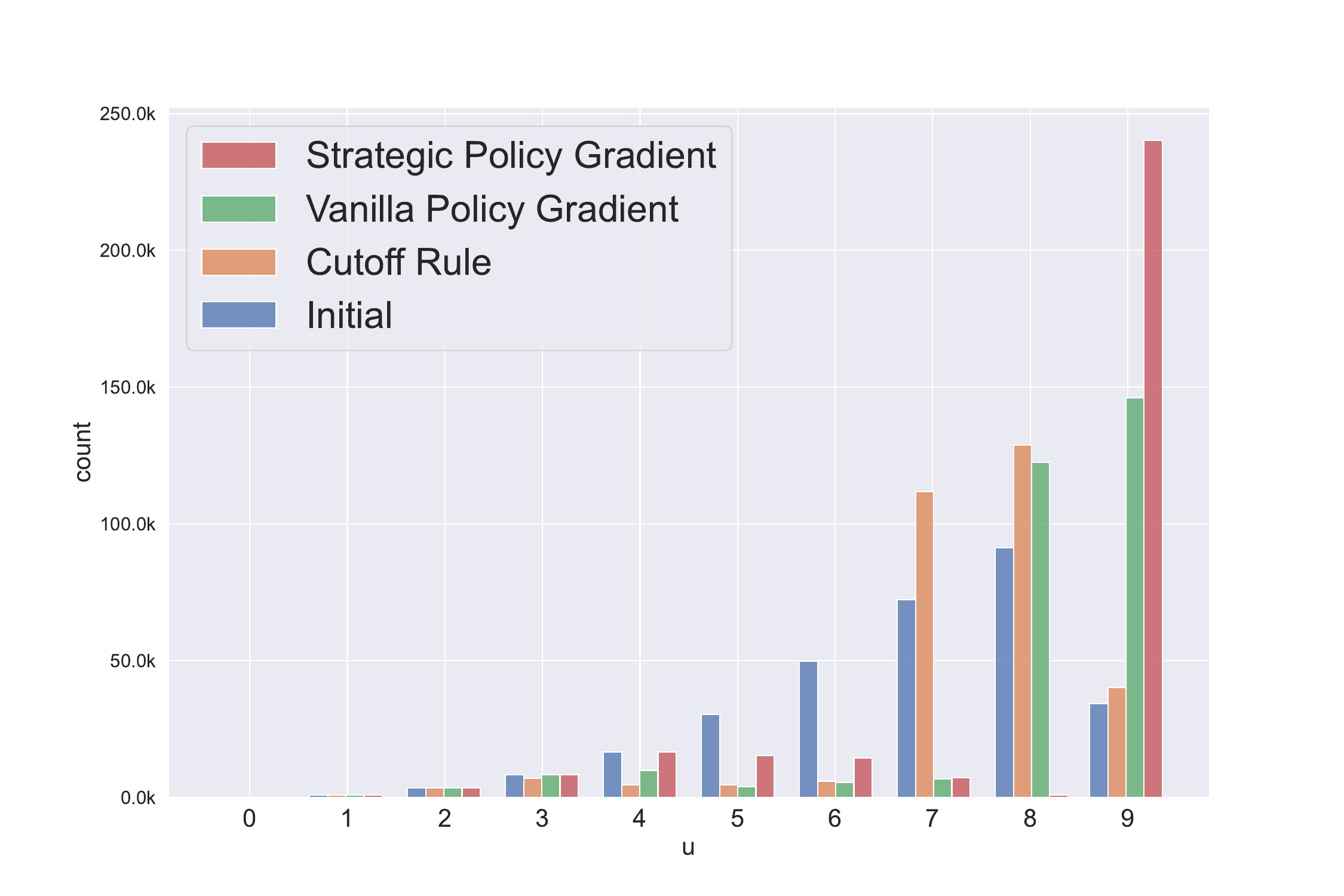}
    \caption{}
    \label{fig:semi_u_distribution_c0.2b}
\end{subfigure}
\caption{The left figure presents the curve of policy value in semi-synthetic experiment with cost coefficient $c=0.2$, and the right figure presents the original distribution of $u$ and that after manipulation induced by policies.}
\label{fig:semi_synthetic_c0.2}
\end{figure}

We conduct the experiments with $c=0.1, 0.15, 0.2$, as cases with lower manipulation costs have already been examined in the synthetic experiment. For illustration, we present the results for $c = 0.1$ and $c = 0.2$ in Figures~\ref{fig:semi_synthetic} and~\ref{fig:semi_synthetic_c0.2}, respectively. The corresponding numerical results are provided in Tables~\ref{table:semi_synthetic_main_paper} and~\ref{table:semi_synthetic_main_paper_c2}. Detailed results can be found in Appendix~\ref{app:semi-syns}.

In the left panel of Figure~\ref{fig:semi_synthetic}, we report the policy value trajectories across 10 successive random seeds. We first observe the dominant performance of our proposed method and its rapid convergence after the warm-up stage, which aligns with the results in the synthetic experiment. A notable phenomenon is that \textit{Vanilla Policy Gradient} outperforms the \textit{Cutoff Rule}, suggesting that the optimal static policy may not remain a performatively stable solution in more complex scenarios.

To provide further insight, the right panel of Figure~\ref{fig:semi_synthetic} and Table~\ref{table:semi_synthetic_main_paper} display the distribution of $u$ before and after manipulation induced by the three methods. Recall that $u$ represents the external credit levels, where higher values correspond to a lower probability of loan default. The original binary label is highly imbalanced, with most agents not defaulting, which is reflected in the skewed distribution of the original $u$. We observe that \textit{Vanilla Policy Gradient} induces more agents to increase their $u$ compared to the \textit{Cutoff Rule}, confirming the advantage of a performativity-aware method, even without explicitly modeling performativity. Remarkably, our proposed method successfully incentivizes all agents to improve their $u$ to the maximum extent, setting it apart in this high-dimensional setting.

\begin{table}[h]
\caption{Experiment result on semi-synthetic data with $c=0.1$}
\label{table:semi_synthetic_main_paper}
\begin{center}
\begin{small}
\begin{sc}
\begin{tabular}{lcccc}
\toprule
Method & $\times 10^5$ Policy value & $\times 10^5$ Policy value & \%Change & Move \\
 & (Best) & (Final epoch) &  &  \\
\midrule
Cutoff & $ 1.74 \pm 0.00$ & $1.74 \pm 0.00$ & 34.41 & 0.79 \\
Vanilla & $2.61 \pm 0.05$ & $2.35 \pm 0.04$ & 49.19 & 1.19\\
Strategic& $3.52 \pm 0.00$ & $3.51 \pm 0.02$ & 88.83 & 2.18\\
\bottomrule
\end{tabular}
\end{sc}
\end{small}
\end{center}
\vskip -0.1in
\end{table}

\begin{table}[h]
\caption{Experiment result on semi-synthetic data with $c=0.2$}
\label{table:semi_synthetic_main_paper_c2}
\vskip 0.15in
\begin{center}
\begin{small}
\begin{sc}
\begin{tabular}{lcccc}
\toprule
Method & $\times 10^5$ Policy value & $\times 10^5$ Policy value & \%Change & Move \\
 & (Best) & (Final epoch) &  &  \\
\midrule
Cutoff & $1.71 \pm 0.00$ & $1.71 \pm 0.00$ & 31.17 & 0.61 \\
Vanilla & $2.53 \pm 0.08$ & $2.26 \pm 0.01$ & 61.79 & 1.25\\
Strategic& $3.13 \pm 0.11$ & $3.08 \pm 0.09$ & 66.97 & 1.26\\
\bottomrule
\end{tabular}
\end{sc}
\end{small}
\end{center}
\vskip -0.1in
\end{table}

We further examine the case of $c=0.2$, where the optimal policy indeed changes, and strategic behavior weakens significantly. Nonetheless, our method still achieves remarkable performance. Moreover, we observe that \textit{Vanilla Policy Gradient} approaches the performance of our method, particularly in the \%CHANGE and MOVE metrics. However, as discussed in the synthetic experiments, these metrics should be considered only as indirect indicators for understanding the policy value.

\section{Conclusion}

In this paper, we address the problem of performative policy learning, situating it within the broader context of decision-making under distribution shifts. Specifically, we interpret performativity as an endogenous shift arising from agents' strategic responses to the deployed algorithm. To overcome key limitations in the existing literature, we propose a methodology characterized by the following core features:
\begin{itemize}
    \item No parametric assumption on utility or data distribution
    \item Limited manipulation, in the light of bounded rationality
    \item Causal mechanism that supports the model to serve as intervention ($\mathcal{D}(f_\theta)$ versus $\mathcal{D}(\theta)$)
    \item Use of batch feedback (versus bandit feedback)
    \item Steer around the micro-level decision process and directly model the decision outcome
    \item Target high-dimensional model parameters and data
\end{itemize}

To enable gradient-based policy optimization in practical scenarios involving high-dimensional data and model parameters, we leverage domain knowledge at both micro and macro levels. This approach facilitates substantial dimensionality reduction, eliminating the need for strong parametric assumptions.

When the manipulatable feature is discrete with relatively low cardinality, which is not only a reasonable assumption but also can be incentivized by a piecewise constant policy that we can design, we propose to leverage the policy evaluation vector as a powerful mediator in the causal path from model to data distribution. We then model the agent behavior by a differentiable classifier, which is trained with the observed manipulatable feature as supervision. For optimizing the policy, we introduce the strategic policy gradient algorithm, supported by a solid convergence guarantee. Additionally, we provide empirical insights and implementation details to enhance practical performance. Challenging experiments further validate the effectiveness of the proposed method.

We argue that high-dimensional model parameters and data represent a practical yet under-explored scenario in performative learning. The basic formulation of performative prediction naturally poses severe computational challenges. We contend that identifying and utilizing reasonable structural information to enhance the practicality of performative learning is both an important and compelling research direction.

\newpage

\bibliographystyle{plain}

\bibliography{ref}



\clearpage
\onecolumn
\appendix

\section{Experiment Details for Synthetic Data}\label{app:synthetic}


\subsection{Parameter setting}\label{app:params_syn}
In this subsection we provide the detailed parameter setting of synthetic experiment, including data generation and policy learning. 
\begin{enumerate}
    \item Dimension of features: $\dim \mathcal{V} = 20$, $|\mathcal{U}|= 5$.
    \item Dimension of (policy function) parameter $\theta$: 1,351
    \item Batch size $n=3000$
    \item Size of validation samples $n_{eval} = 10000$. The samples used for policy evaluation (for generating charts) is fixed across the whole session.
    \item Time horizon $T = 100$
    \item Warm-up duration $T_0 = 30$
    \item Weight vector $\beta$ in CATE expression: For $u$ in $\{0,1,2,3,4\}$, we set weight sequence that linearly located in interval $[-5,5]$. For every dimension of $v$, we generate the weight from standard Gaussian distribution.
    \item Weight matrix $W$ in generating $u\mid v$: for every term in the matrix, we generate it from $\mathcal{N}(1,1)$. This matrix is fixed through all experiments.
    \item Policy network $\pi_\theta$: 2-layer MLP with ReLU activation and Sigmoid output. Hidden layer size: $2(\dim \mathcal{V} + |\mathcal{U}|)=50$. 
    \item Behavior network $h_\gamma$: 3-layer MLP with ReLU activation. Hidden layer size: $2(\dim \mathcal{V} + |\mathcal{U}|)=50$.
    \item Optimizer of $\pi_\theta$: Adam~\cite{kingma2017adammethodstochasticoptimization} with learning rate $0.05$ for warm-up stage, and Adagrad~\cite{JMLR:v12:duchi11a} with learning rate $0.05$ after warm-up stage.
    \item Optimizer of $h_\gamma$: SGD with learning rate $0.1$.

\end{enumerate}

We comment that the dimension of $\theta$ is greater than 1,000, which is very high-dimensional, and break the setting in former literature, where the dimension is usually lower than 10.

\subsection{Baselines}\label{app:baseline}


In this subsection, we explain the three baselines in detail.

The \textit{Cutoff Rule} is the optimal one in the environment of no strategic behavior, which is implemented using the \textbf{oracle} CATE in synthetic experiment and semi-synthetic experiment. 
 

The \textit{Vanilla Policy Gradient} applies the idea of repeated gradient descent \cite{perdomo2020performative}, though gradient ascent is used here to maximize policy value. This baseline represents the method that is performativity-aware but does not model performativity explicitly. 

The \textit{End2end Policy Gradient} uses the same pattern with our proposed strategic policy gradient, including warm-up stage, and early-stop in the training of behavior model. The only difference lies in the way of modeling behavior $p(u\mid v, \pi_\theta)$. In \textit{End2end Policy Gradient}, behavior model takes $\theta$ and $v$ as input to predict the post-manipulation $u$, contrasting $\zeta(\theta,v)$ and $v$ acting as input in strategic policy gradient. This baseline represents the method that model the performativity through the idea of distribution map $\mathcal{D}(\theta)$, and help to validate the benefit of model (prediction) as intervention.

Additionally, in all of gradient-based methods, we adopt linear regression in synthetic experiment and XGBoost \cite{chen2016xgboost} in semi-synthetic experiment to estimate the CATE $\tau(x)$ and plug such estimate into the gradient expression.

Then, we explain why we do not consider zeroth-order optimization (refer to \cite{flaxman2005online, miller2021outside}). One of the core settings of this paper is high-dimensional model parameter $\theta$, which may come from modern neural network. While such zeroth-order optimization belongs to methods that employ bandit-feedback, which is extremely sample-inefficient since it compresses the batch samples into single feedback, as also discussed in \cite{miller2021outside, jagadeesan2021alternative}, not to mention the objective function could be non-convex.

Furthermore, we do not implement \textit{End2end Policy Gradient} in the semi-synthetic experiment, since the number of parameter of policy network is much larger than that in synthetic case, and this method scales badly. Moreover, we have checked and analyzed the performance of this baseline in synthetic experiment in detail.

\subsection{Experiment results and strategic behavior}\label{app:result}
Since different manipulation cost coefficient $c$ corresponds to different manipulation patterns, we present the result under four different levels of $c\in \{0.05, 0.1, 0.15\}$ as tables \ref{table:synthetic_0.05} to \ref{table:synthetic_0.1}.

In each table, we report the policy value of the best in the whole trajectory and that in the final epoch. We also calculate the proportion of improving $u$ among agents and the mean change (with sign) of $u$ induced by different policies learned by different methods, in which the average is calculated among all agents. Though not reported here, we comment that the proportion of manipulation is nearly equal to the proportion of improvement for all methods, which means all of them avoid inducing negative manipulation. Taking this into consideration, one can deduce the mean move among agents that indeed move through column MOVE dividing column \%CHANGE. 

We observe that the induced manipulation in our proposed method becomes weaker when the manipulation cost coefficient $c$ increases. This is reflected by the decrease of both the proportion of improvement and mean move. However, we can calculate that the mean move among those who indeed improve is nearly unchanged when $c$ increases, which means the agents owning the potential to improve are still well-incentivized.

We also observe that the strategic behavior induced by the other three baselines is insensitive to the cost coefficient here. Since the cutoff policy, CATE$>0$ is the performative stable point of these two methods, we can explain this phenomenon through the mechanism of the cutoff policy. Under this policy, only agents whose CATE is just below the borderline $0$ would improve their $u$, and agent whose CATE is either positive or negative too far from $0$ has no incentive to improve their $u$. We call the agents with CATE just below the borderline as potential improvers, and the size of this group will decrease only for a much larger cost coefficient. To illustrate it, we provide the following policy value curve for cutoff policy under different manipulation cost coefficients $c$. 
\begin{figure}[h]
    \centering
\includegraphics[width=0.7\columnwidth]{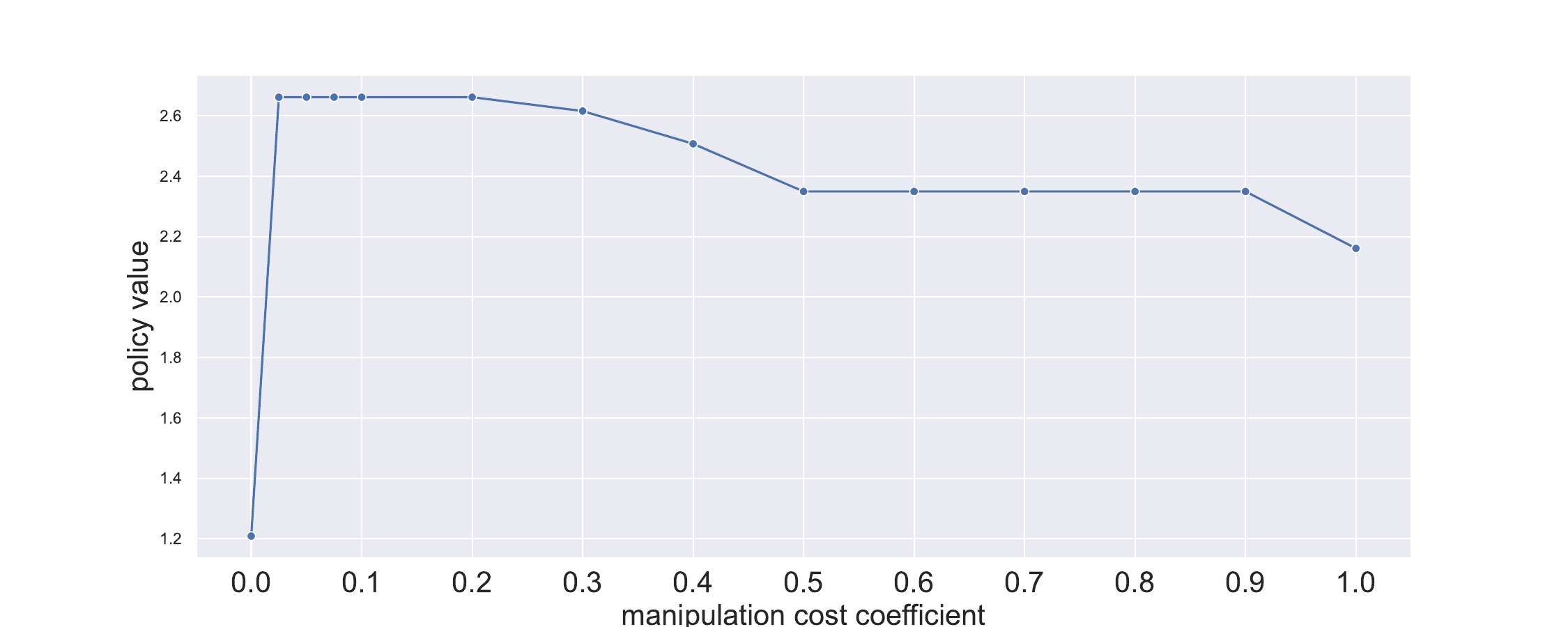}
    \caption{The policy value of cutoff policy under different manipulation cost coefficient $c$.}
    \label{fig:cutoff_cc}
\end{figure}
In addition, a good policy not only induce improvement but also allocate treatment correctly according to estimated CATE. For instance, we still should not assign high propensity to a loan applicant with a very bad credit profile (characterized by fixed feature $v$), even if he has the potential to improve. To be concrete, the effect on inducing improvement of setting the propensities as $0.7,0.8$ and/or $0.1,0.2$ for two $u$ levels is the same, but the policy value can differ a lot. Thus, these two indices used for characterizing the invoked strategic behavior can only be a side witness on the policy value.

\begin{table}[!ht]
\caption{Experiment result on synthetic data with \textbf{$c=0.05$}}
\label{table:synthetic_0.05}
\vskip 0.15in
\begin{center}
\begin{small}
\begin{sc}
\begin{tabular}{lcccc}
\toprule
Method & Policy value (Best) & Policy value (Final epoch) & \%Change & Move \\
\midrule
Cutoff & $2.92 \pm 0.00$ & $2.92 \pm 0.00$ & 37.50 & 0.74\\
Vanilla & $2.89 \pm 0.02$ & $2.86 \pm 0.02$ & 30.24& 0.59\\
End2end & $2.62 \pm 0.02$ & $2.55 \pm 0.07$ & 20.73 & 0.41 \\
Strategic& $5.22 \pm 0.03$ & $5.19 \pm 0.04$ & 76.54 & 1.92\\
\bottomrule
\end{tabular}
\end{sc}
\end{small}
\end{center}
\vskip -0.1in
\end{table}

\begin{table}[!ht]
\caption{Experiment result on synthetic data with \textbf{$c=0.1$}}
\label{table:synthetic_0.75}
\vskip 0.15in
\begin{center}
\begin{small}
\begin{sc}
\begin{tabular}{lcccc}
\toprule
Method & Policy value (Best) & Policy value (Final epoch) & \%Change & Move \\
\midrule
Cutoff & $2.92 \pm 0.00$ & $2.92 \pm 0.00$ & 37.50 & 0.74\\
Vanilla& $2.96 \pm 0.02$ & $2.89 \pm 0.01$ & 30.82& 0.59\\
End2end & $2.76 \pm 0.06$ & $2.76 \pm 0.06$ &24.29 & 0.48 \\
Strategic& $5.26 \pm 0.03$ & $5.23 \pm 0.03$ & 74.38 & 1.87\\
\bottomrule
\end{tabular}
\end{sc}
\end{small}
\end{center}
\vskip -0.1in
\end{table}

\begin{table}[!ht]
\caption{Experiment result on synthetic data with \textbf{$c=0.15$}}
\label{table:synthetic_0.1}
\vskip 0.15in
\begin{center}
\begin{small}
\begin{sc}
\begin{tabular}{lcccc}
\toprule
Method & Policy value (Best) & Policy value (Final epoch) & \%Change & Move \\
\midrule
Cutoff & $2.92 \pm 0.00$ & $2.92 \pm 0.00$ & 37.50 & 0.74\\
Vanilla& $2.99 \pm 0.02$ & $2.92 \pm 0.01$ & 31.12 & 0.59\\
End2end& $2.86 \pm 0.06$ & $2.86 \pm 0.06$ & 26.20 & 0.51 \\
Strategic& $5.25 \pm 0.06$ & $5.21 \pm 0.07$ & 70.88 & 1.80\\
\bottomrule
\end{tabular}
\end{sc}
\end{small}
\end{center}
\vskip -0.1in
\end{table}

\subsection{Instances of training behavior model}\label{app:impact_h}

\begin{figure}[ht]
\centering
\begin{subfloat}
    \centering
    \includegraphics[width=0.8\columnwidth]{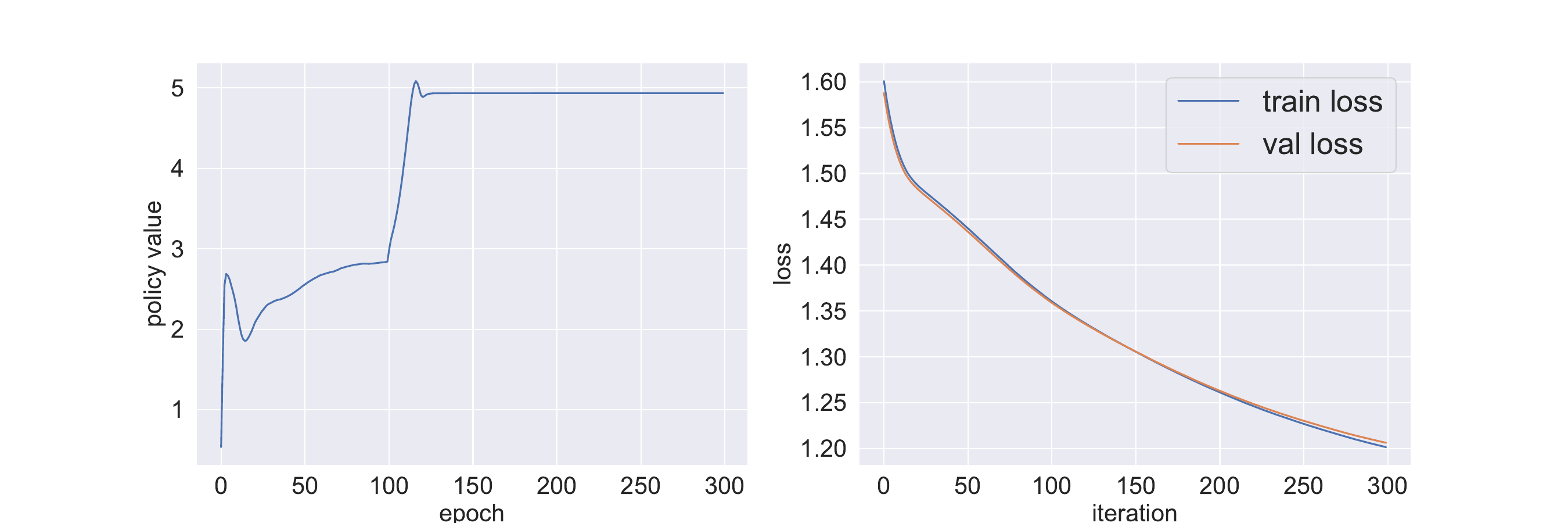}
\end{subfloat}
\begin{subfloat}
    \centering
    \includegraphics[width=0.8\columnwidth]{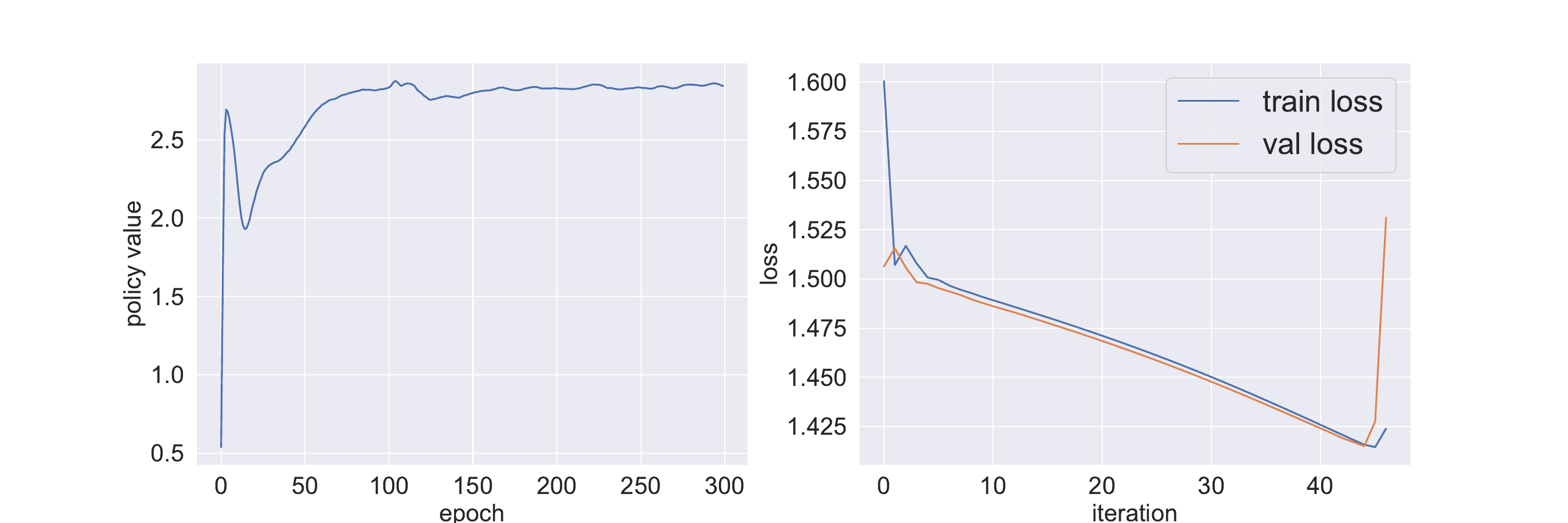}
\end{subfloat}
\caption{The upper figure is the training curve of proposed strategic policy gradient, with the upper left policy value curve and upper right loss curve. The lower figure is that of end2end policy gradient.
}\label{fig:stra_vs_e2e}
\end{figure}

In this subsection, we supplement some instances of training behavior models. We illustrate that the way to model the behavior matters, for which we show the training curve of strategic policy gradient and end2end policy gradient in Figure \ref{fig:stra_vs_e2e}. 

It should be mentioned once more that the difference between our proposed strategic policy gradient and end2end policy gradient only lies in the way to model and approximate the agent behavior, i.e. $p(u \mid v, \pi_\theta)$. We can observe that the learning of behavior model in end2end policy gradient behaves badly, which echoes our analysis on the drawback of $\theta$ as mediator in the main paper. Since early stopping is adopted in the training of behavior model $h_\gamma$, we can also infer that the number of effective training epochs in end2end policy is much less than that of strategic policy gradient. 

\newpage

\section{Experiment Details for Semi-synthetic Data}\label{app:semi-syns}

\subsection{Parameter Setting}

In this subsection, we provide the detailed parameter setting  in the semi-synthetic experiment:
\begin{enumerate}
    \item Dimension of features: $\dim \mathcal{V} = 55$, $|\mathcal{U}|= 10$.
    \item Dimension of (policy function) parameter $\theta$: 25,741
    \item Number of total rounds: $T=300$.
    \item Number of warm-up rounds: $T_0=100$.
    \item Policy network $\pi_\theta$: $3$-layer MLP with ReLU activation and Sigmoid output. Hidden layer size: $2(\dim \mathcal{V} + |\mathcal{U}|)=130$.
    \item Behavior network $h_\gamma$: $3$-layer MLP with ReLU activation and Softmax output. Hidden layer size: $2(\dim \mathcal{V} + |\mathcal{U}|)=130$.
    \item Optimizer of $\pi_\theta$: Adam with learning rate $0.01$ for warm-up stage, and Adagrad with learning rate $0.01$ after warm-up stage.
    \item Optimizer of $h_\gamma$: SGD with learning rate $0.01$.
\end{enumerate}

Since the degree of non-linearity of objective function increases greatly in semi-synthetic contrasting synthetic case, we adopt Adam as optimizer with momentum. Nevertheless, as mentioned in the main paper, high-proportional momentum will make our methods incur instability, as well as vanilla gradient. Thus, we select a lower level of $(\beta_1,\beta_2)$ compared to $(0.9, 0.999)$ in traditional case.

Additionally, we mention that the training of behavior model $h_\gamma$ is based on a random $80\%$-$20\%$ training/validation set split, and loss on validation set is used for early-stopping. Moreover, the policy value at every epoch is evaluated using the entire dataset, which is a bit different from synthetic experiment. It is noteworthy that the evaluation of policy value does not influence the update of policy and is only utilized for illustration.

\subsection{Experiment Results}

The experimental results with semi-synthetic data are presented in table \ref{table:semi_synthetic_policy_value_c1} and \ref{table:semi_synthetic_policy_value_c2}. We report both the best policy value during all epochs and the policy value at the final epoch. Additionally, we report the proportion of agents whose $u$ are improved induced by policies, and the mean change (with sign) of $u$ induced by policies. The table \ref{table:semi_synthetic_policy_value_c1}  is a complementary of Figure \ref{fig:semi_synthetic}. 

In the semi-synthetic experiment, we consider the cost coefficient $c=0.1,0.15,0.2$. Since we've observed that our proposed strategic policy gradient can persuade all agents to improve $u$ to the highest level, we do not repeat the experiments with $c<0.1$ here. An important phenomenon is that our methods oscillate more when $c$ increases to $0.2$, with the total policy value decreasing naturally. We attribute this phenomenon to the increase of hardness of learning behavior model $h_\gamma$, as more agents would not improve his $u$, which may blur the gradient estimate $\nabla_\theta q(u\mid v,\pi_\theta)$. As mentioned in the main paper, we also observe that our method will further degenerate into the vanilla gradient for larger $c$, e.g. $c=0.3$ (though it is very high for the best response mechanism in Equation (\ref{eq:best_response_semi})).

At last, we discuss the estimation of CATE. It is noteworthy that the CATE structure is much more complex than the synthetic case, and we also check the performance of the following method: learning CATE with constant treatment allocation (e.g. $\pi(x)=0.9$) and committing the CATE$>0$ with estimated CATE instead of the oracular one. The average policy value of this method is around $1.29$ (though not presented in detail here), which accords with our expectation. We mention this to elicit the complicated interplay of policy learning and correspondingly generated data. Policy generates the data, which is used for the update of policy in turn. This is another complex topic concerned with endogeneity, not to mention the complexity of the performative distribution shift combined with it. Based on the logic of pure-exploration, it is intuitive that a good policy update process should not only target policy value, but also consider generating the needed data adaptively for its further update, and the constant allocation policy above is a counterexample. Though all these gradient-based methods do not implement this idea explicitly, it may be done implicitly, and we think the internal mechanism of gradient-based methods in performative policy learning is worthy of further exploration.






\begin{table}[h]
\caption{Experiment result on semi-synthetic data with $c=0.1$}
\label{table:semi_synthetic_policy_value_c1}
\vskip 0.15in
\begin{center}
\begin{small}
\begin{sc}
\begin{tabular}{lcccc}
\toprule
Method & $\times 10^5$ Policy value & $\times 10^5$ Policy value & \%Change & Move \\
 & (Best) & (Final epoch) &  &  \\
\midrule
Cutoff & $ 1.74 \pm 0.00$ & $1.74 \pm 0.00$ & 34.41 & 0.79 \\
Vanilla & $2.61 \pm 0.05$ & $2.35 \pm 0.04$ & 50.17 & 1.19\\
Strategic& $3.52 \pm 0.00$ & $3.51 \pm 0.02$ & 88.83 & 2.18\\
\bottomrule
\end{tabular}
\end{sc}
\end{small}
\end{center}
\vskip -0.1in
\end{table}

\begin{table}[h]
\caption{Experiment result on semi-synthetic data with $c=0.15$}
\label{table:semi_synthetic_policy_value_c15}
\vskip 0.15in
\begin{center}
\begin{small}
\begin{sc}
\begin{tabular}{lcccc}
\toprule
Method & $\times 10^5$ Policy value & $\times 10^5$ Policy value & \%Change & Move \\
 & (Best) & (Final epoch) &  &  \\
\midrule
Cutoff & $1.74 \pm 0.00$ & $1.74 \pm 0.00$ & 34.07 & 0.76 \\
Vanilla & $2.50 \pm 0.04$ & $2.28 \pm 0.08$ & 51.64 & 1.20\\
Strategic& $3.50 \pm 0.03$ & $3.48 \pm 0.08$ & 87.40 & 2.08\\
\bottomrule
\end{tabular}
\end{sc}
\end{small}
\end{center}
\vskip -0.1in
\end{table}

\begin{table}[h]
\caption{Experiment result on semi-synthetic data with $c=0.2$}
\label{table:semi_synthetic_policy_value_c2}
\vskip 0.15in
\begin{center}
\begin{small}
\begin{sc}
\begin{tabular}{lcccc}
\toprule
Method & $\times 10^5$ Policy value & $\times 10^5$ Policy value & \%Change & Move \\
 & (Best) & (Final epoch) &  &  \\
\midrule
Cutoff & $1.71 \pm 0.00$ & $1.71 \pm 0.00$ & 31.17 & 0.61 \\
Vanilla & $2.53 \pm 0.08$ & $2.26 \pm 0.01$ & 61.79 & 1.25\\
Strategic& $3.13 \pm 0.11$ & $3.08 \pm 0.09$ & 66.97 & 1.26\\
\bottomrule
\end{tabular}
\end{sc}
\end{small}
\end{center}
\vskip -0.1in
\end{table}

\section{Proofs}
\subsection{Proof of Proposition \ref{prop1}}\label{app:proof_of_prop1}

\textbf{Proof}. Note that 
\begin{equation}
    \begin{aligned}
p(x_i;\pi_{\theta}) &= p(u_i,v_i;\pi_\theta)  &\phantom{}\\
&= p(u_i|v_i,\pi_\theta)p(v_i|\pi_\theta) &\phantom{}\\
&= p(u_i|v_i,\pi_\theta)p(v_i) & (\pi_\theta \text{ only influence } u_i)\\
&=p(u_i|v_i, [\pi_{\theta}((u,v_i))]_{u\in \mathcal{U}}) p(v_i)  & (\pi_\theta \text{ is summarized as evaluations}  )\\ 
&=p(u_i|[\pi_{\theta}((u,v_i))]_{u\in \mathcal{U}}) p(v_i)  & (\text{conditional independence})
\end{aligned}
\end{equation}

Thus, 
\begin{equation}
    \nabla_\theta\log p(x_i;\pi_{\theta}) = \nabla_\theta\log p(u_i|v_i, [\pi_{\theta}((u,v_i))]_{u\in \mathcal{U}})
\end{equation}

Next, we apply the chain rule and get:
\begin{equation}
    \nabla_\theta \log p\left(x ; \pi_\theta\right)=\nabla_\zeta \log p(u \mid \zeta) \nabla_\theta \zeta(v, \pi_\theta)
\end{equation}
which is the desired result. \hfill \qedsymbol

\subsection{Proof of Lemma~\ref{lemma1}}\label{app:proof_of_lemma1}

\textbf{Proof}. For the existence of gradient, we refer to the Theorem 1 in~\cite{zhou2008derivative}. Now we proceed given the existence of gradients. The spectral representation tells us that for $f\in \mathcal{H}$, we have:
\begin{equation}
    f(\zeta)=\langle f, K(\zeta, \cdot)\rangle_{\mathcal{H}}
\end{equation}
Thus, given differentiability, we have:
\begin{equation}
    \nabla_\zeta f(\zeta)=\left\langle f, \nabla_\zeta K(\zeta, \cdot)\right\rangle_{\mathcal{H}}
\end{equation}
Here, for each dimension $i$ of vector $\zeta$, $\nabla_\zeta K(\zeta, \cdot)$ still lies in the same RKHS $\mathcal{H}$. Thus, the RHS above actually denotes following column vector:
\begin{equation}
    \left\langle f, \nabla_\zeta K(\zeta, \cdot)\right\rangle_{\mathcal{H}} = (\left\langle f, \nabla_{\zeta_i} K(\zeta, \cdot)\right\rangle_{\mathcal{H}})_{i=1}^{|\mathcal{U}|}
\end{equation}

Thus, by Cauchy-Schwarz inequality, we have:
\begin{equation}
    \|\nabla_\zeta f(\zeta)-\nabla_\zeta g(\zeta)\|=\left\|\left\langle f-g, \nabla_\zeta K(\zeta, \cdot)\right\rangle_{\mathcal{H}}\right\| 
    \leq
    \|f-g\|_{\mathcal{H}} \cdot\left\|\left(\left\|\nabla_{\zeta_i} K(\zeta, \cdot)\right\|_{\mathcal{H}}\right)_{i=1}^{|\mathcal{U}|}\right\|
\end{equation}
which finishes the proof. \hfill \qedsymbol

\subsection{Proof of Theorem~\ref{theorem1}}\label{app:proof_of_theorem1}

\textbf{Proof}. 
As a preview, we will first prove that our gradient estimate
\begin{equation}
    \hat g
    = \frac{1}{n}\sum_{i=1}^n    
    \nabla_\theta \pi_\theta\left(x_i\right) \hat{\tau}\left(x_i\right)+\pi_\theta\left(x_i\right) \hat{\tau}\left(x_i\right) \nabla_\theta \log q\left(u_i \mid \zeta(v_i, \pi_\theta)\right)
\end{equation}
can approximate the true performative gradient
\begin{equation}
    \nabla_\theta V(\pi_\theta) 
    = \mathbb{E}\left[
    \nabla_\theta \pi_\theta\left(X\right) {\tau}\left(X\right)+\pi_\theta\left(X\right) {\tau}\left(X\right) \nabla_\theta \log p\left(u \mid \zeta(v, \pi_\theta)\right)
    \right]
\end{equation}
with any precision. Here, we use $X$ to stress the randomness of feature vector, while slightly abusing the notation $u,v$ (instead of $U,V$) for neatness. When there is an outer expectation, $X=(u,v)$ contains randomness, whereas $u,v$ also denotes single points of manipulatable/fixed feature when point-wise bound is applied and no expectation is involved.

We clarify that the expectation above is taken with respect to the data distribution $X \sim \mathcal{D}(\pi_{\theta_t})$ in each epoch $t$, which occurs after the warm-up stage. For simplicity, we set the time step $t = 1$ as the beginning of using the performative gradient and denote the distribution at time $t$ as $\mathcal{D}_t = \mathcal{D}(\pi_{\theta_t})$.

We further denote the data distribution used to train the behavior model and CATE estimator as $\mathcal{D}_0$, which can be a mixture distribution, such as the average of $\mathcal{D}(\pi_{\theta_0})$ for $\theta_0 \in \Theta_0 \subset \Theta$, where $\Theta_0$ represents a set of random samples corresponding to random initialization of model parameter. We assume this i.i.d. structure to avoid the issue of adaptivity in the training data, which is too complex to address and beyond the scope of this paper. This distribution, $\mathcal{D}_0$, will be used when discussing the estimation error of $q(\cdot \mid \zeta)$.

Now, let us return to the main storyline. First, we need a concentration inequality to transform $\hat g$ into its expected version, which we denote as:
\begin{equation}
    \nabla_\theta\hat V(\pi_\theta)  = 
    \mathbb{E}\left[
    \nabla_\theta \pi_\theta\left(X\right) {\hat\tau}\left(X\right)+\pi_\theta\left(X\right) {\hat\tau}\left(X\right) \nabla_\theta \log q\left(u \mid \zeta(v, \pi_\theta)\right)
    \right]
\end{equation}

Since we have assumed the boundedness of $\nabla_\theta \pi_\theta(x)$, $\tau(x)$, and $\nabla_\theta q\left(u \mid \zeta(v, \pi_\theta)\right)$, we can apply Hoeffding's inequality to construct the required concentration, which yields, with probability at least $1 - \delta_1$,
\begin{equation}
    \|\hat g - \nabla_\theta\hat V(\pi_\theta)\|  = O\left(\frac{\log(1/\delta_1)}{n}\right)
\end{equation}

Next, we can focus on the gap:
\begin{equation}
    \begin{aligned}
        \|
        \nabla_\theta \hat V(\pi_\theta)
        -  
        \nabla_\theta V(\pi_\theta)     
        \| 
        &= 
        \left\|
        \mathbb{E}\left[
        \nabla_\theta \pi_\theta\left(X\right) 
        \left(\hat\tau(X) - \tau(X)\right)
        +    
        \pi_\theta\left(X\right) \left(\hat\tau(X)-\tau(X) \right)
        \nabla_\theta \log q\left(u \mid \zeta(v, \pi_\theta)\right)
        \right.    
        \right. \\
        &\phantom{+}
        \left. \left.
        + \pi_\theta(X)\tau(X) 
        \left(
        \nabla_\theta \log q\left(u \mid \zeta(v, \pi_\theta)\right)
         - \nabla_\theta \log p\left(u \mid \zeta(v, \pi_\theta)\right)
        \right)
        \right]
        \right\|
\end{aligned}    
\end{equation}

The first two terms can be easily handled through the bound on the estimation error of the CATE. We focus on the third term, specifically the distance between the logarithm of the true performative gradient and that of its estimate. Given Proposition~\ref{prop1}, we apply the chain rule and obtain:
\begin{equation}
    \nabla_\theta \log p(u \mid \zeta(v, \pi_\theta)
=
\nabla_\zeta \log p(u \mid \zeta) \nabla_\theta \zeta(v, \pi_\theta)
\end{equation}
and 
\begin{equation}
    \nabla_\theta \log q(u \mid \zeta(v, \pi_\theta)=\nabla_\zeta \log q(u \mid \zeta) \nabla_\theta \zeta(v, \pi_\theta)
\end{equation}
Since the knowledge on $\pi_\theta$ is exact, there is no estimation error on $\nabla_\theta \zeta(v, \pi_\theta)$. Hence, our core step is bounding:
\begin{equation}
    \left\|
    \mathbb{E}\left[
    \nabla_\zeta \log q(u\mid \zeta) - \nabla_\zeta \log p(u\mid \zeta) 
    \right]
    \right\|
\end{equation}

By Jensen's inequality, we have:
\begin{equation}
    \left\|
    \mathbb{E}\left[
    \nabla_\zeta \log q(u\mid \zeta) - \nabla_\zeta \log p(u\mid \zeta) 
    \right]
    \right\| \leq 
    \mathbb{E}\left[
    \left\|
    \nabla_\zeta \log q(u\mid \zeta) - \nabla_\zeta \log p(u\mid \zeta) 
    \right\|
    \right]    
\end{equation}

For simplicity, we temporally hide the variable $\zeta$ and write $p(u \mid \zeta)$ as $p$ herein. The following argument applies to each $u \in \mathcal{U}$. Moreover, the gradient is taken with respect to $\zeta$ by default. Notice that:
\begin{equation}
    \begin{aligned}
\|\nabla  \log p - \nabla \log q\|
&= 
\|
\frac{\nabla p}{p} - \frac{\nabla q}{q} 
\|
\\
&= 
\|
\frac{q\nabla p - p \nabla q}{pq} 
\|
\\
&=
\|
\frac{q\nabla p - p\nabla p + p\nabla p -p \nabla q}{pq} 
\|
\\
&\leq 
\|
\frac{(q-p)}{qp} \nabla p
\|
+
\|
\frac{1}{q} (\nabla p - \nabla q)
\|
\\
&\leq 
\left(
\frac{G_p}{\iota^2} + \frac{G_K\sqrt{|\mathcal{U}|}}{\iota}
\right) | p - q|
\end{aligned}
\end{equation}

We denote:
\begin{equation}
    \epsilon_t = \mathbb{E}[| p(u \mid \zeta) - q(u \mid \zeta)|]
\end{equation}
where the expectation is taken with respect to $(u^\prime,v)\sim \mathcal{D}_t$. 

An important observation is that, although there are distribution shifts for each $t \geq 1$ compared to $\mathcal{D}_0$, which is used to train the estimator $q$, the lower bound $\iota$ on the p.m.f. ensures that $\mathcal{D}_t$ is absolutely continuous with respect to each other. Thus, all $\epsilon_t$ share the same convergence rate with respect to $n$, and we denote:
\begin{equation}
    \epsilon = \sup_t \epsilon_t   
\end{equation}
Here, the supremum over $t$ corresponds to all possible distributions $\mathcal{D}_t$.

Concerning the convergence of $\epsilon$, we can construct the following lemma, which demonstrates that this estimation error can indeed converges to 0 with high probability. For clarity, we defer the details of this lemma to the end of this proof.
\begin{lemma}\label{lemma:convergence_of_estimation_error}
Under necessary regularity assumptions, we have the following relation holds with probability at least $1-\delta_2$:
    \begin{equation}
    \mathbb{E}[|p-q|] = O\left(n^{-1/4}\left(\sqrt{\|p\|_\mathcal{H}/\iota} + \sqrt[4]{\log(1/\delta_2)}\right)\right)
\end{equation}
\end{lemma}

Now, let us return to the main storyline. On the other hand, we can easily bound $|\nabla_\theta \zeta(v, \pi_\theta)|$ as follows:
\begin{equation}
    \|\nabla_\theta \zeta(v, \pi_\theta)\| \leq G_\pi\sqrt{|\mathcal{U}|}
\end{equation}

Combining these results, we can bound the gap between the log gradients as follows:
\begin{equation}
    \begin{aligned}
&\|
\nabla_\theta \log p(u \mid \zeta(v, \pi_\theta)) 
- 
\nabla_\theta \log q(u \mid \zeta(v, \pi_\theta))
\|
\\
\leq& 
\|\nabla  \log p - \nabla \log q\| \cdot 
\|\nabla_\theta \zeta(v, \pi_\theta)\|
\\
\leq &
\epsilon \left(\frac{G_p}{\iota^2} + \frac{G_K\sqrt{|\mathcal{U}|}}{\iota} \right) G_\pi\sqrt{|\mathcal{U}|}
\end{aligned}
\end{equation}

Now let's go forward to the gap between $\nabla_\theta V(\pi_\theta),\nabla_\theta \hat V(\pi_\theta)$. We have:
\begin{equation}
\begin{aligned}
\|\nabla_\theta V(\pi_\theta) - \nabla_\theta \hat V(\pi_\theta)\|
&=
\left\|
\mathbb{E}\left[
\nabla_\theta \pi_\theta\left(X\right) \left(\hat{\tau}(X) - \tau(X)\right) 
+
\pi_\theta(X)\left(
\tau(X) \nabla_\theta \log p(u \mid \zeta(v, \pi_\theta)
\right. \right.\right.
\\
&\phantom{=} - 
\left.\left.\left.
\hat\tau(X) \nabla_\theta \log q(u \mid \zeta(v, \theta))
\right)
\right.
\right]
\|
\\
&= 
\left\|\mathbb{E}\left[
\nabla_\theta \pi_\theta\left(X\right) \left(\hat{\tau}(X) - \tau(X)\right) 
+
\pi_\theta(X) \left(
(\tau(X)-\hat\tau(X))\nabla_\theta\log p(u \mid \zeta(v, \pi_\theta) \right.\right.\right.
\\
&\phantom{=} + 
\left.\left.\left. 
\hat\tau(X) \left(
\nabla_\theta\log p(u \mid \zeta(v, \pi_\theta) - \nabla_\theta\log q(u \mid \zeta(v, \pi_\theta)
\right)
\right)
\right]
\right\|
\\
&\leq 
\frac{\epsilon_\tau G_\pi}{\iota} +  \frac{\epsilon_\tau G_p G_\pi \sqrt{\mathcal{U}}}{\iota}  + \epsilon B_\tau \left(\frac{G_p}{\iota^2} + \frac{G_K\sqrt{|\mathcal{U}|}}{\iota} \right) G_\pi\sqrt{|\mathcal{U}|}
\end{aligned}
\end{equation}
Here, we use importance weights to transition the estimation error of CATE from $\mathcal{D}_t$ to $\mathcal{D}_0$:
\begin{equation}
    \begin{aligned}
\left|\mathbb{E}_{X\sim \mathcal{D}_t}[\tau(X) - \hat\tau(X)] \right|
&= 
\left|\mathbb{E}_{X\sim \mathcal{D}_0}[\frac{p_t(X)}{p_0(X)}\left(\tau(X) - \hat\tau(X)\right)]\right|
\\
&= 
\left|\mathbb{E}_{X\sim \mathcal{D}_0}[\frac{p_t(u\mid v)}{p_0(u\mid v)}\left(\tau(X) - \hat\tau(X)\right)]\right|
\\
&\leq
\left|
\frac{1}{\iota}\mathbb{E}_{X\sim \mathcal{D}_0}[\left(\tau(X) - \hat\tau(X)\right)]\right|
\end{aligned}
\end{equation}
and the following upper bound on the logarithm of performative density.
\begin{equation}
    \begin{aligned}
\|\nabla_\theta \log p(u \mid \zeta(v, \pi_\theta)) \|  
&= 
\|
\frac{\nabla_\zeta p(u \mid \zeta)}{p(u \mid \zeta) }
\nabla_\theta \zeta(v, \pi_\theta)
\|
\\
&\leq 
\frac{G_p G_\pi\sqrt{\mathcal{U}}}{\iota} 
\end{aligned}
\end{equation}

Therefore, as the estimation errors $\epsilon_p, \epsilon_\tau$ approach 0, we have $|\nabla_\theta V(\pi_\theta) - \nabla_\theta \hat{V}(\pi_\theta)|$ also approaching 0.

For simplicity, we denote:
\begin{equation}
    V_t = V(\pi_{\theta_t}) \quad \hat V_t = \hat V(\pi_{\theta_t})
\end{equation}
and
\begin{equation}
    E_t =\nabla_\theta \hat V_t - \nabla_\theta V_t
\end{equation}

Now we can delineate the convergence through standard procedures in convex optimization.

The concavity and $l$-smoothness of the policy value $V$ gives the standard bound with second-order Taylor expansion: 
\begin{equation}
    V_{t+1} \geq V_t + (\nabla_\theta V_t)^\top (\theta_{t+1}-\theta_t) - \frac{l}{2}\|\theta_{t+1} - \theta_t\|^2
\end{equation}
Then we plug-in the parameter update $\theta_{t+1} = \theta_t + \eta g_t$ and $g_t = \nabla_\theta \hat V_t+\kappa\sqrt{\log (1/\delta_1)/n}$, here $\kappa$ is a constant. 
\begin{equation}
    V_{t+1} \geq V_t + \eta(\nabla_\theta V_t)^\top (\nabla_\theta \hat V_t) - \frac{l\eta^2}{2}\left(\|\nabla_\theta \hat V_t\|^2 + \frac{\kappa^2\log (1/\delta_1)}{n}\right)
\end{equation}
which is equivalent to:
\begin{equation}
    V_{t+1} \geq V_t + \eta\left(\|\nabla_\theta  V_t\|^2  + (\nabla_\theta V_t)^\top E_t \right) - \frac{l\eta^2}{2}\|\nabla_\theta V_t + E_t \|^2 
    - 
    \frac{l\eta^2\kappa^2\log (1/\delta_1)}{2n}
\end{equation}

Now we rearrange this inequality:
\begin{equation}
    V_{t+1} - V_t + (l\eta^2-\eta)(\nabla_\theta V_t)^\top  E_t + \frac{l\eta^2}{2}\|E_t\|^2 
    \geq 
(\eta-\frac{l\eta^2}{2})\|\nabla_\theta V_t\|^2
- 
    \frac{l\eta^2\kappa^2\log (1/\delta_1)}{2n}
\end{equation}

By Cauchy-Schwarz inequality, we know that:
\begin{equation}
    |(\nabla_\theta V_t)^\top  E_t| \leq \|\nabla_\theta V_t\|\cdot \|E_t\|
\end{equation}

Notice that we can bound $\|\nabla_\theta V_t\|$ as following:
\begin{equation}
    \begin{aligned}
\|\nabla_\theta V(\pi_\theta) \|
&= 
\left\|\nabla_\theta \pi_\theta\left(x\right) {\tau}\left(x\right)+\pi_\theta\left(x\right) {\tau}\left(x\right) \nabla_\theta \log p\left(u \mid \zeta(v, \pi_\theta)\right)\right\|
\\
&\leq
G_\pi B_\tau + B_\tau \frac{G_p G_\pi\sqrt{\mathcal{U}}}{\iota} 
\end{aligned}
\end{equation}

For simplicty, we now denote
\begin{equation}
    G_V = G_\pi B_\tau + B_\tau \frac{G_p G_\pi\sqrt{\mathcal{U}}}{\iota}
\end{equation}
as the upper bound on $\|\nabla_\theta V(\pi_\theta) \|$, and 
\begin{equation}
    G_E = \frac{\epsilon_\tau G_\pi}{\iota} + \frac{\epsilon_\tau G_p G_\pi \sqrt{\mathcal{U}}}{\iota}  + \epsilon B_\tau \left(\frac{G_p}{\iota^2} + \frac{G_K\sqrt{|\mathcal{U}|}}{\iota} \right) G_\pi\sqrt{|\mathcal{U}|}
\end{equation}
as the upper bound on $\|E_t\|$, which has been proved. 

To proceed, we now have:
\begin{equation}
    V_{t+1} - V_t + |l\eta^2-\eta|G_VG_E + \frac{l\eta^2}{2} G_E^2
\geq 
(\eta - \frac{l\eta^2}{2})\|\nabla_\theta V_t\|^2
- 
    \frac{l\eta^2C^2\log (1/\delta_1)}{2n}
\end{equation}
To make the inequality above meaningful, we should make the learning rate satisfy:
\begin{equation}
    \eta - \frac{l\eta^2}{2} > 0
\end{equation}
which is equivalent to:
\begin{equation}
    \eta < \frac{2}{l}
\end{equation}

Notice that the inequality above holds for $t=1,2,...,T$ (we only implement one step of gradient ascent in every epoch), with probability at least $1-T\delta_1\delta_2$. We sum them and get:
\begin{equation}
    \sum_{t=1}^T \|\nabla_\theta V_t\|^2 
\leq 
\frac{V_{T+1} - V_1 + T|l\eta^2-\eta|G_VG_E  +\frac{Tl\eta^2}{2} G_E^2 +  
    \frac{Tl\eta^2C^2\log (1/\delta_1)}{2n}}
{(\eta-\frac{l\eta^2}{2})}
\end{equation}
For clarification, we mention that $t=1$ corresponds to the start of applying the estimated performative gradient, wherein the warm-up stage has ended. 

We now know that:
\begin{equation}
    \min_{1\leq t\leq T} \|\nabla_\theta V_t\|^2 
\leq  
\frac{1}{T}\sum_{t=1}^T \|\nabla_\theta V_t\|^2 
\leq 
\frac{\frac{1}{T}(V_{T+1} - V_1) + |l\eta^2-\eta|G_VG_E  +\frac{l\eta^2}{2} G_E^2
+ 
    \frac{l\eta^2C^2\log (1/\delta_1)}{2n}}
{(\eta-\frac{l\eta^2}{2})}
\end{equation}

Notice that we can bound the policy value by $B_\tau$:
\begin{equation}
    |V\left(\pi_\theta\right)|=|\mathbb{E}_{x \sim p\left(; \pi_\theta\right)}\left[\pi_\theta(x) \tau(x)\right]| \leq B_\tau 
\end{equation}

Thus, we get the final bound:
\begin{equation}
    \min_{1\leq t\leq T} \|\nabla_\theta V_t\|^2  \leq \frac{\frac{2}{T}B_\tau + |l\eta^2-\eta|G_VG_E  +\frac{l\eta^2}{2} G_E^2 + 
    \frac{l\eta^2C^2\log (1/\delta_1)}{2n}
    }
{(\eta-\frac{l\eta^2}{2})}
\end{equation}

Since we have proved that $G_E$ approaches to 0 with the estimation error of CATE, $\epsilon_\tau$ and that of performative distribution, $\epsilon$, approaching to 0, we finish the proof of Theorem~\ref{theorem1}.  \hfill \qedsymbol

\vskip 1em

\textbf{\textit{Proof of Lemma~\ref{lemma:convergence_of_estimation_error}}.} Using the standard results from \cite{bach2024learning}, specifically Proposition 4.5, Equations (7.13) and (7.14), we have:
\begin{equation}\label{eq:risk_error_bound}
\mathbb{E}\left[\mathcal{R}\left(q\right)\right]-\mathcal{R}\left(p\right) 
\leq 
G \sqrt{2 \inf _{q \in \mathcal{H}}\left\{\left\|q-p\right\|_{L_2(\mathcal{D}_0)}^2+\frac{16 R^2}{n}\|p\|_{\mathcal{H}}^2\right\}} 
\end{equation}

Here, $\mathcal{R}$ is the expected risk function:
\begin{equation}
    \mathcal{R}(f)=\mathbb{E}[\ell(u^\prime, f(\zeta))]
\end{equation}
with $\ell$ denoting the cross-entropy loss, and the expectation in the risk is taken with respect to $(u', v) \sim \mathcal{D}_0$.

In addition, the expectation on the $\mathcal{R}(q)$ is taken with respect to the randomness of $n$ samples drawn from $\mathcal{D}_0$ for training $q$.

Since the $\inf$ on the RHS of Equation (\ref{eq:risk_error_bound}) is taken over the entire RKHS, and under the realizability assumption, i.e., $p \in \mathcal{H}$, we can derive an upper bound on the RHS by setting $q = p$, which gives:
\begin{equation}
\mathbb{E}\left[\mathcal{R}\left(q\right)\right]-\mathcal{R}\left(p\right) 
\leq 
\frac{4\sqrt{2}GR}{\sqrt{n}} \|p\|_\mathcal{H}
\end{equation}

Here, $G$ is the Lipschitz constant of the loss function. Since we impose a lower bound $\iota$ on both $p$ and $q$, we directly obtain that $G = 1/\iota$. $R$ represents the upper bound of the feature map in the corresponding RKHS, which is a constant determined by the kernel function.

Next, again, by concentration inequality with necessary regularity assumptions, we can also transform the expected risk into risk with a specific $q$, with probability $1-\delta_2$:
\begin{equation}
        \mathcal{R}\left(q\right)-\mathcal{R}\left(p\right) 
= 
O\left(\frac{\|p\|_\mathcal{H}/\iota + \sqrt{\log(1/\delta_2)}}{\sqrt{n}}\right)
\end{equation}

We then plug-in the cross-entropy loss, which gives:
\begin{equation}
    \mathcal{R}\left(q\right)-\mathcal{R}\left(p\right)  = 
    \mathbb{E}_v\left[ 
    D_{KL}\left(p(\cdot \mid \zeta) \| q(\cdot \mid \zeta) \right) \right]
\end{equation}
The original expectation of the risk is taken with respect to $(u', v) \sim \mathcal{D}_0$. We apply the law of iterated expectations and first condition on $v$, which corresponds to the inner expectation taken with respect to $p(u \mid \zeta(v, \pi_\theta))$. This leads to the KL divergence between the two discrete (conditional) distributions, $p(\cdot \mid \zeta)$ and $q(\cdot \mid \zeta)$.

By Pinsker's inequality, we can transform the KL divergence into its lower bound, the total variation distance:
\begin{equation}
    D_{TV}\left(p(\cdot \mid \zeta) \| q(\cdot \mid \zeta) \right) 
    \leq 
    \sqrt{
    \frac{
    D_{KL}\left(p(\cdot \mid \zeta) \| q(\cdot \mid \zeta) \right)
    }
    {2}    
    }
\end{equation}
where $D_{TV}$ is given by:
\begin{equation}
    D_{TV}\left(p(\cdot \mid \zeta),q(\cdot \mid \zeta)\right)=\frac{1}{2} \sum_u|p(u \mid \zeta)-q(u \mid \zeta)|
\end{equation}

Putting things together, we get:
\begin{equation}
    \mathbb{E}_{\mathcal{D}_0} [|p-q|] = O\left(n^{-1/4}\left(\sqrt{\|p\|_\mathcal{H}/\iota} + \sqrt[4]{\log(1/\delta_2)}\right)\right)
\end{equation}

Notice that the distribution of the fixed feature $v$ remains constant, and the lower bound $\iota$ on the p.m.f. ensures the absolute continuity of $\mathcal{D}_t$ with respect to $\mathcal{D}_0$. As a result, all $\epsilon_t$ share the same convergence rate. We conclude that:
\begin{equation}
    \epsilon = O\left(n^{-1/4}\left(\sqrt{\|p\|_\mathcal{H}/\iota} + \sqrt[4]{\log(1/\delta_2)}\right)\right)
\end{equation}
\hfill \qedsymbol

\subsection{Proof of Proposition \ref{prop2}}\label{app:proof_of_prop2}

\textbf{Proof}. We prove it by contradiction. We assume that an agent manipulate its feature $u$ to $u^\prime \in (a_j,a_{j+1})$. First, we can conclude that $u\notin (a_j,a_{j+1})$, since otherwise $\pi(u,v)=\pi(u^\prime,v)$, and thus $u\succ u^\prime$, i.e. agent will strictly prefer to keep unchanged.

Next, we should only consider the case that $u,u^\prime$ locates in different intervals $I$. When $u > u^\prime$ holds, we have $(a_{j+1}-\epsilon_u) \succ u^\prime$ for any $\epsilon_u \in (0, a_{j+1}-u^\prime)$, i.e. $a_{j+1}-\epsilon_u$ is a a preferred option for sufficiently small $\epsilon_u$, when it approaches $a_{j+1}$ sufficiently. Similarly, when $u < u^\prime$ holds, we have $a_j \succ u^\prime$, which also make contradiction. Since each feature has a minimal measurement scale, denoted by $\epsilon_u$, we argue that agent will move to a point just to the left of a knot, with a gap of $\epsilon_u$.

Thus, we conclude that if agent decides to manipulation, i.e. there exists point $u^\prime \succ u$, then $u^\prime$ either locates at one of the knots or approaches a knot infinitely.
\hfill \qedsymbol

\end{document}